\documentclass[12pt]{amsproc}

\usepackage[
  left=2.5cm,
  right=2.5cm,
  top=2cm,
  bottom=3cm
]{geometry}

\usepackage{orcidlink}
\usepackage{todonotes}
\usepackage{listings}
\usepackage{booktabs}
\usepackage{makecell}
\usepackage{doi}
\usepackage[final]{changes}
\usepackage{hyperref}
\newcommand{\rreff}{\textnormal{ref}}
\newcommand{\FLOPs}{\textnormal{FLOPs}}

\usepackage[accsupp]{axessibility}
\newenvironment{aiuse}{%
  \par\vspace{11pt}%
  \noindent\textbf{Declaration on the use of AI:}\enspace}{%
  \par\vspace{11pt}}

\newenvironment{data}{%
  \par\vspace{11pt}%
  \noindent\textbf{Data availability:}\enspace}{%
  \par\vspace{11pt}}

\newenvironment{acknowledge}{%
  \par\vspace{11pt}%
  \noindent\textbf{Acknowledgements:}\enspace}{%
  \par\vspace{11pt}}

\title[Algorithmic algorithm development with LLMs]{Algorithmic algorithm development with LLMs: \\ A Case Study on LLM-Usage for Contraction Order Optimization in Tensor Networks} 
\author[F. Hoppe, M. Röhrig-Zöllner, P. Knechtges]{Fabian Hoppe, Melven Röhrig-Zöllner, Philipp Knechtges \vspace{6pt}\\  \itshape German Aerospace Center (DLR), Institute of Software Technology, \\ High-Performance Computing, Cologne (Germany)} 


\begin{document} 
\maketitle 

\begin{abstract}We consider LLM-based algorithm development through a case study on contraction-order optimisation for tensor networks with OpenEvolve. We pay particular attention to the choice of the LLM as well as design choices such as evaluation metric and test instances. Our results highlight both the promise of verifier-guided evolutionary coding agents for algorithm development/improvement and the continuing importance of evaluation, validation, and interpretation---and corresponding challenges---by the human scientist.
\end{abstract}


\section{Introduction}

As in many areas, LLM-based systems are increasingly used in science in recent times. In addition to their widespread use for coding assistance and auxiliary tasks such as literature research, summarisation, etc., the potential of AI for core scientific tasks is increasingly being taken into account. End-to-end ``AI scientist'' systems \cite{lu_ai_scientist_arxiv_2024,yamada_ai_scientist_v2_arxiv_2025,gottweis2025coscientist,mitchener2025kosmosaiscientistautonomous} aim to automate larger parts of the research cycle, including idea generation, coding and experimentation, and manuscript preparation---up to now, however, their capabilities appear to be limited, as critical assessments show \cite{beel_ai_scientist_reproduced_2025}. In this paper we take a step back and consider LLM usage for a core scientific task, namely algorithm development and/or improvement, but without end-to-end, i.e., idea-to-paper, automation. We put particular emphasis on the design decisions and assumptions that the human scientist must provide to the respective AI system. 

Regarding contributions to coding assistance tailored to science, we mention exemplarily the SciCode benchmark \cite{tian_scicode_arxiv_2024} and agentic approaches to scientific computing such as Re4 \cite{cheng_re4_arxiv_2025} and JutulGPT \cite{lie_agentic_scientific_simulation_arxiv_2026}. During the last two years, several approaches for LLM-based algorithm engineering, often also overlapping with performance engineering, searching for heuristics, or even broader applicability, have been presented. In the broader context we mention, e.g., ``Evolution of Heuristics'' (EoH, \cite{liu_eoh_arxiv_2024}), \cite{ye_reevo_arxiv_2024}, RedAHD \cite{thach_redahd_arxiv_2025}, \cite{gl_ns_arxiv_2026}, and \cite{surina_algorithm_discovery_rl_arxiv_2025}. More closely to the topic investigated in the present paper are FunSearch \cite{romeraparedes2024funsearch}, AlphaEvolve \cite{novikov_alphaevolve_arxiv_2025}, OpenEvolve, GigaEvo \cite{khrulkov2025gigaevoopensourceoptimization}, and CodeEvolve \cite{assumpcao2025codeevolve}, as well as ShinkaEvolve \cite{lange2025shinkaevolve}, DeepEvolve \cite{liu2025scientificalgorithmdiscoveryaugmenting}, LLM4AD \cite{liu2024llm4ad}, ThetaEvolve \cite{wang2025thetaevolve}, Escher-Loop \cite{liu2026escherloopmutualevolutionclosedloop}, and AlgoTuner \cite{press_algotune_arxiv_2025}. Roughly speaking, EoH, ReEvo, RedAHD, and G-LNS mainly address automated heuristic design for combinatorial optimization, whereas FunSearch and the AlphaEvolve/OpenEvolve line search directly over executable program variants under automated evaluation; LLM4AD, in turn, is best viewed as a unifying platform rather than one specific search strategy. For systematic evaluation of such systems, benchmarks like ALE-Bench \cite{imajuku_alebench_arxiv_2025} or the AlgoTune benchmark \cite{press_algotune_arxiv_2025} have been introduced; ALE-Bench focuses on long-horizon score-based optimization tasks, whereas AlgoTune evaluates speedups on expert-provided numerical programs under correctness and runtime constraints. Often, however, an ad-hoc selection of problems from various areas is used. Applications include, but are not limited to, e.g., mathematical discovery \cite{georgiev2025mathematicalexplorationdiscoveryscale}, systems performance research \cite{cheng2025barbarians}, automated discovery of language model scaling laws \cite{lin2026languagemodelsdiscoverscaling}, finding new acquisition functions in Bayesian optimization \cite{aglietti2024funbodiscoveringacquisitionfunctions}, and performance optimization \cite{nagaitsev2025optimizingpytorchinferencellmbased}. 

At the same time, first challenges with the use of LLM-based systems in scientific software development are emerging. As large quantities of software can increasingly be produced automatically without major effort, authorship and appreciation may increasingly be shifting towards high-level design, verification by humans and the associated assumption of responsibility, as well as transparency and sustainability; the JOSS governance documents\footnote{see, e.g., \url{https://joss.readthedocs.io/en/latest/submitting.html}, \url{https://blog.joss.theoj.org/2026/01/preparing-joss-for-a-generative-ai-future} [Accessed 05/2026]} can be interpreted as an indication for this shift. Furthermore, there are concerns regarding stability \cite{rajput_dynamic_stability_arxiv_2025}, reproducibility \cite{siddiq_reproducibility_crisis_arxiv_2025}, digital sovereignty (especially for sensitive areas of research), and impacts on the social, organizational, and career aspects of research software. In fact, even our concept of what software engineering actually is, could be changing, e.g., by extending the object of engineering from executable code to ``semi-executable artifacts'' \cite{feldt2026semiexecutablestackagenticsoftware} such as prompts, workflows, evaluation harnesses, controls, and organizational routines. Also, there are similar, albeit more philosophical, considerations on a possible change in the self-image of mathematical-scientific work due to AI \cite{klowden2026mathematicalmethodshumanthought}. Closer to the present scope is recent work on human-AI-interaction for mathematical discovery with AlphaEvolve \cite{baeuerle2026}: it is argued that this process is not a one-off question-answer interaction, but rather an iterative process of ``intentmaking''---defining and sharpening the goal, setting up and updating the experiment---and ``sensemaking''---the interpretation, validation and debugging of the results. Consequently, it is suggested that such kind of system should be viewed and used as collaborative scientific instruments rather than as question-answer assistants. 

The aim of this publication is not to tell a ``success story'' for one or more specific problems. Instead, we want to use a manageable, but non-trivial and realistic example to analyse the factors that contribute to success or failure. In other words, given a concrete algorithmic problem and a concrete LLM-based system for algorithm search, we consider, as an example, the design of the experiment to be provided by the human scientist and analyse its effects as systematically as possible. Specifically, we use OpenEvolve to determine a contraction sequence with low FLOP count for a special class of tensor networks. \added{As will be explained in more detail in Section \ref{subsec:tn} below, the problem of determining a contraction order with as few FLOPs as possible for a given tensor network is difficult to solve, but easy to verify: for a given contraction order, the number of FLOPs can be calculated without the actual execution of the contraction and at the same time one can check whether the given order is permissible. In fact, it is sufficient to know only the shape of all tensors involved, whereas ``real'' tensors available in memory are not necessary. This suggests a strong conceptual fit to what OpenEvolve can do: evolving code on the basis of a reliable verification/scoring routine---for more details see Section \ref{subsec:openevolve} below. The fact that during the evaluation only the candidate program (i.e. contraction order determination) but not the contraction itself needs to be evaluated on the respective test examples seems to suggest an unusually cheap setting for computing scores. Moreover, by avoiding actual performance measurements that would be hardware dependent and noisy, the overall setting becomes more controllable. This enables us, in particular, to carry out the experiments in Python;  nevertheless, the overall approach applies in principle to other programming languages supported by the underlying LLM and situations that require actual performance measurements, too.} 

\added{Related but distinct work has recently used LLMs for tensor-network \emph{structure} search rather than contraction ordering \cite{zeng2024tngps,iacovides2025domainaware}. In the broader context of quantum computing, GigaEvo has recently been applied to algorithm improvement for quantum compilation \cite{fisher2026llmguidedevolutionarysearchalgebraic}. To the best of our knowledge, however, there are no publications that apply LLM-based algorithm search specifically to the problem of tensor network contraction ordering yet. 
Even though our test problem seems to be very well suited for OpenEvolve and unusually benign, the situation is by no means trivial, as we will experimentally show in the following sections.}
We examine the influence of the (open) LLMs used, the sample data used during optimization and the metric used for optimization, and then analyse the ``best'' result found in detail.
In view of the large number of different approaches and the extremely rapid development of the entire environment, the results presented here, which are limited to one specific approach, OpenEvolve, are necessarily limited, of course. Nevertheless, we believe that our detailed observations and considerations on the influencing factors in automatic algorithm search are relevant and can be generalised to similar problems at least in the broadest sense. Moreover, given the rapid progress in this field, the snapshot presented here may one day be of ``historical'' interest.

\paragraph*{Structure of the Paper.} In Section \ref{sec::background} we provide the reader with the necessary background information on OpenEvolve and the problem of tensor network contraction. The overall experimental setup is described at the beginning of Section \ref{sec::experiments}, followed by the \added{main} results of different experiments and an in-depth analysis of the ``best'' result. Concluding considerations and lessons learned are stated in Section \ref{sec::conclusion} \added{and additional technical details and results are provided in the Appendix.}

\section{Background}\label{sec::background}

\subsection{OpenEvolve} 
\label{subsec:openevolve}

OpenEvolve\footnote{\url{https://github.com/algorithmicsuperintelligence/openevolve} [Accessed 05/2026]. We use version 0.2.25.} can be understood as an \emph{evolutionary coding agent}. It is an open-source Python-based framework inspired by DeepMind's closed-source AlphaEvolve \cite{novikov_alphaevolve_arxiv_2025}, which itself builds on ideas from the earlier FunSearch approach \cite{romeraparedes2024funsearch}. While FunSearch searches in the space of implementations of a designated function inside a fixed scaffold, AlphaEvolve/ OpenEvolve generalise this idea to direct edits of larger code artefacts. In its current implementation, OpenEvolve combines LLM-generated code mutations with island-based evolutionary search and a MAP-Elites-style quality-diversity archive. The user needs to provide (i) an initial program and (ii) an automated evaluator; OpenEvolve then repeatedly asks one or more LLMs to propose edits, executes the resulting candidates, scores them with the evaluator, and retains or recombines promising variants through evolutionary search. Hence, this approach is suitable for problems with a reliable automatic verifier and a measurable objective. For algorithmic and implementation details we refer the reader to the repository, and only note the practically relevant detail that OpenEvolve allows the explicit propagation of random seeds across major components to improve reproducibility. 

Among the applications this approach let us highlight just a few: The original AlphaEvolve white paper reports improvements for circle packing problems and a new multiplication algorithm for $4 \times 4$ matrices, but also more practical applications such as data-center scheduling \cite{novikov_alphaevolve_arxiv_2025}; further applications, including production level ones, have recently been highlighted in a blog post\footnote{\url{https://deepmind.google/blog/alphaevolve-impact/} [Accessed 05/2026]}. OpenEvolve has so far been used for, e.g., combinatorial bijection discovery \cite{brown2025bijection} and systems-oriented AI-driven research \cite{cheng2025barbarians}, multi-agent coordination \cite{kumar2026evolvinginterpretableconstitutionsmultiagent}, password guessing \cite{mazin2026llmguidedpromptevolutionpassword}, technology mapping \cite{fu2026mappingevolve}, and improved estimates for certain mathematical constants \cite{bhan2026newboundszarankiewicznumbers}. 

Being based on population-based verifier-guided search, OpenEvolve differs from other agentic approaches in the field. Compared to budgeted iterative edit-run-select agents such as AlgoTuner \cite{press_algotune_arxiv_2025}, it does not primarily follow a single incumbent solution but maintains a population of competing candidate programs and lets external task metrics drive selection. Relative to prompt-evolution approaches such as Promptbreeder or GEPA, it mutates and selects \emph{program source code} rather than prompts or agent policies \cite{fernando2023promptbreeder,agrawal2025gepa}. Different from multi-agent scientific assistants such as AI co-scientist \cite{gottweis2025coscientist}, which aim to cover larger parts of the scientific cycle, OpenEvolve performs closed-loop optimization in a program space defined by the user-provided verifier/evaluator. Among the evolutionary coding frameworks in the field, OpenEvolve is by construction closest to AlphaEvolve, whereas CodeEvolve \cite{assumpcao2025codeevolve} and GigaEvo \cite{khrulkov2025gigaevoopensourceoptimization} put more emphasis on the orchestration of the evolutionary search, LLM4AD \cite{liu2024llm4ad} provides a broader experimental platform for LLM-based algorithm development, ShinkaEvolve  \cite{lange2025shinkaevolve} focuses strongly on sample-efficient evolution via parent/LLM selection and novelty rejection-sampling, and ThetaEvolve \cite{wang2025thetaevolve} extends the approach towards in-context and reinforcement learning at test time. Escher-Loop \cite{liu2026escherloopmutualevolutionclosedloop} extends OpenEvolve towards a self-improving evolution system. Finally, \cite{liu2025scientificalgorithmdiscoveryaugmenting} combines OpenEvolve with Deep Research in order to incorporate knowledge beyond the LLMs internal knowledge.

\subsection{Tensor Network Contraction} 
\label{subsec:tn}

\begin{figure}[t]
    \centering
    \includegraphics[width=0.25\columnwidth]{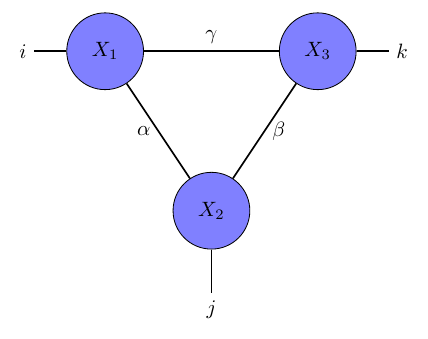}
    \includegraphics[width=0.25\columnwidth]{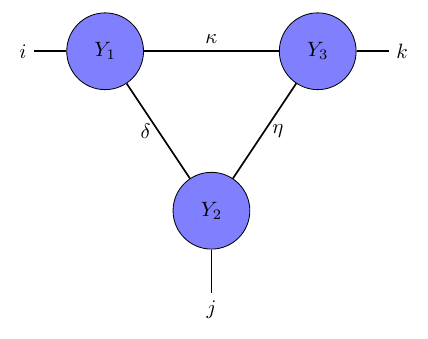}
    \includegraphics[width=0.3\columnwidth]{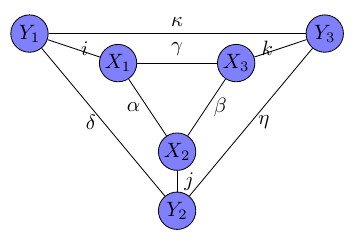}
    \caption{Examples for tensor networks}
    \label{fig:tn} 
\end{figure}

For the purpose of this paper, it suffices to consider a tensor as a multidimensional array with entries indexed by a finite set of \emph{indices} (sometimes also called ``modes'' or ``legs''). For example, a rank-$3$ tensor $A$ with indices $(i,j,k)$ has entries $A_{ijk}$, whereas a rank-$0$ tensor is just a number. A \emph{tensor network} (TN) is a collection of tensors together with an identification of certain indices between tensors, indicating that those indices are to be \emph{summed over}, a procedure referred to as ``contraction''. TNs are usually described by graphs (or hypergraphs): vertices represent tensors, and edges represent shared indices. Contracting all shared indices yields either a scalar (if no indices remain) or a tensor (if indices, so-called ``free indices'', remain). For an introduction we refer the interested reader to, e.g., \cite{Orus2014PracticalIntro,BiamonteBergholm2017Nutshell,Ballard2025}.
 
As an example let us consider two three-dimensional tensors $X$ and $Y$ with free indices $i,j,k$, both represented as so-called periodic Matrix Product States (MPS, \cite{mps2004}); cf. the two graphs on the left in Figure \ref{fig:tn}. $X$ is represented as contraction of $X_1,X_2,X_3$ over the indices $\alpha,\beta,\gamma$, and similarly for $Y$. Contracting now $X$ and $Y$ along the shared indices $i,j,k$, which is basically a generalised scalar product between $X$ and $Y$, results in the tensor network on the right-hand side of Figure \ref{fig:tn}.  
\begin{align*}
&\sum_{i,j,k} X[i,j,k] Y[i,j,k] \\ &= \sum_{i,j,k} \left( \sum_{\alpha,\beta,\gamma} X_1[\gamma,i,\alpha]\,
X_2[\alpha,j,\beta]\,
X_3[\beta,k,\gamma]\, \right) \left( \sum_{\delta,\eta,\kappa} Y_1[\kappa,i,\delta]\,
Y_2[\delta,j,\eta]\,
Y_3[\eta,k,\kappa] \right) \\ &=
\sum_{i,j,k,\alpha,\beta,\gamma,\delta,\eta,\kappa}
X_1[\gamma,i,\alpha]\,
X_2[\alpha,j,\beta]\,
X_3[\beta,k,\gamma]\,
Y_1[\kappa,i,\delta]\,
Y_2[\delta,j,\eta]\,
Y_3[\eta,k,\kappa]
\end{align*}
Contracting two tensors is a direct generalization of matrix multiplication and can be realized numerically by repeated reshaping and matrix multiplication (GEMM-like kernel) operations. Here, the order in which the indices are contracted is crucial, as computational complexity and also memory requirements for intermediate results can vary dramatically with respect to the chosen order \cite{Orus2014PracticalIntro,OGorman2019Parameterization}. Apparently this is also the case for our example in Figure \ref{fig:tn} (cf. also the formula).
As for matrix multiplication, the number of FLOPs for contracting a TN given a specified contraction order can be computed without performing the actual computation. Finding optimal contraction orders, however, is in general NP-hard \cite{Dumitrescu2018TreewidthBenchmark,Xu2023NPHardness} and thus computationally intractable except for very small instances. Consequently, contraction orders for TNs are often determined using heuristics or meta-heuristics, or by dedicated algorithms for restricted families of networks. We restrict ourselves to a very concise overview over some common approaches:

\begin{itemize}
    \item Greedy and random-greedy heuristics are fast and simple, and can be surprisingly effective on many instances \cite{SmithGray2018opteinsum,GrayKourtis2021hyperoptimized}
    \item Optimization-based: simulated annealing and genetic algorithms \cite{SchindlerJermyn2020Algorithms}
    \item (Hyper-)graph- and tree-based methods \cite{MarkovShi2008,Dumitrescu2018TreewidthBenchmark,GogateDechter2004QuickBB,Strasser2017FlowCutter,Tamaki2017PIDTreewidth,SchlagEtAl2021HighQualityHypergraphPartitioning,StaudtEtAl2024ImprovedCutStrategy,Ibrahim2022ContractionTrees}  
    \item Hybrid frameworks and search over hyperparameters \cite{GrayKourtis2021hyperoptimized,StaudtEtAl2024ImprovedCutStrategy}
    \item AI-driven approaches such as reinforcement learning \cite{MeiromEtAl2022RL,LiuZhang2023RLQuantum}
    \item approaches tailored to very specific structures, e.g., tree tensor networks or tensor-train-type scalar products \cite{Stoian2024TreeTN,Torri2026}.
\end{itemize}

As we will use cotengra\footnote{\url{https://github.com/jcmgray/cotengra}, \url{https://cotengra.readthedocs.io/en/latest/} [Accessed 05/2026]. We use version 0.7.5 for our experiments.}, a Python library implementing one of these approaches, \cite{GrayKourtis2021hyperoptimized}, as baseline in our subsequent experiments, we provide some more details; cotengra is aimed at contraction planning for tensor networks and focuses on the ``planning'' layer, i.e., finding a high-quality contraction tree/path, that then can be executed using standard tensor primitives. In this context, cotengra provides data structures for tensor networks and routines for computing FLOPs, maximal intermediate sizes, etc. for a given contraction order/tree. On an algorithmic level, multiple heuristics (greedy, partitioning-based, local reconfiguration, slicing) are combined with randomized search/hyper-optimization.

\section{Experiments and Results}\label{sec::experiments}

Obviously, the experiments \added{presented} in this Section are only a subset of those one could perform in theory. Our selection may be justified as follows: First, we performed those experiments that were of highest interest to us, namely those influencing variables by means of which the human defines the AI system's room for manoeuvre and defines the goal: the choice of test data and the choice of the metric for evaluation. As there is a large amount of literature on prompting, cf., e.g., \cite[Sect. 5.7]{SurveyLLMsForCode2026}, we do not investigate different system messages. Second, among the hyper-parameters of the agentic system itself, we decided to focus on the choice of the LLM as this potentially raises some very delicate questions; in comparison, we consider the analysis of the remaining hyper-parameters (temperature, top\_p, etc. for the LLM, and population size, number of islands, sampling temperature, etc. for the evolutionary algorithm) to be less interesting---although clearly relevant for performance---and also as too expensive in terms of compute due to the much larger overall parameter space. Third, we address the difficult issue of reproducibility as much as possible within a reasonable budget of compute. Finally, when addressing the quality of the ``best'' outcome, we \added{take a pragmatic approach and use a limited number of non-specialized baselines that are available in a single library;} our point \added{here} is not the actual outcome itself and how good it is (in the sense of an algorithmic contribution), but rather which influencing factors and pitfalls need to be considered on the way there and which questions arise from this. \added{Already focussing on optimizing FLOP counts alone is a very strong design decision, here largely made for the sake of simplicity. For the practical feasibility of a contraction sequence the maximum size of the intermediate is also decisive, and even the runtime is in reality dependent on other influencing variables (e.g., memory speed) that go beyond the pure FLOP count.} 

\added{To keep the presentation brief and clear, we limit ourselves to the essentials in the following; the interested reader is referred to the Appendix for further information and details. We generate three different sets of tensor networks referred to as ``small'', ``middle'' and ``large'' in the following; all networks are of type ``scalar product of two periodic MPS'' but differ in length and dimensions. Reference contraction orders for all networks are obtained with cotengra;} we note that the options referred to as {\itshape ``cotengra cheap''} in the following are described as ``cheap to run [...] but will still yield much better results than simple algorithms'' in the documentation\footnote{\url{https://cotengra.readthedocs.io/en/latest/basics.html\#economical-optimizer} [Accessed 05/2026]}; therefore, we decided to use this as our main baseline. Consequently, the feature dimensions that guide evolution are designed as follows: Given a tensor network $x$, denote by $o_A(x)$ the contraction order for $x$ obtained by algorithm $A$ and by $o_{\rreff}(x)$ the contraction order by the baseline ``cotengra cheap''. Let $\FLOPs(o_A(x),x)$ and $\FLOPs(o_{\rreff}(x),x)$ denote the corresponding number of FLOPs. Then we define the log10\_speedup as   
\begin{align}\label{def:log10speedup}
\textnormal{log10\_speedup}(x,o_A) = \log_{10} \FLOPs(o_{\rreff}(x),x) - \log_{10} \FLOPs(o_A(x),x),
\end{align}
and correspondingly the average (``avg\_log10\_speedup”), median (``median\_log10\_speedup”), maximum (``max\_log10\_speedup”), minimum (``min\_log10\_speedup”) over all tensor networks $x$ contained in the respective test set $\mathcal{T}$. Similarly, we define 
\begin{equation}
\label{def:log10reduction}
\begin{aligned}
\textnormal{log10\_total\_flops\_reduction}
&:= \log_{10}
    \sum_{x \in \mathcal{T}}
    \FLOPs\bigl(o_{\rreff}(x),x\bigr)
\\
&\qquad
    - \log_{10}
    \sum_{x \in \mathcal{T}}
    \FLOPs\bigl(o_A(x),x\bigr).
\end{aligned}
\end{equation}
\added{Metrics defined upon \eqref{def:log10speedup} aggregate the individual, i.e., per-network, improvement in the ``order of magnitude`` of FLOPs over the entire test set, whereas \eqref{def:log10reduction} measures the cumulative improvement in the order of magnitude of FLOPs required for the entire set of TNs.}
To obtain the ``combined\_score'' metric that actually guides evolution, we define 
\begin{equation}
\label{def:completescore}
\textnormal{combined\_score}
:=
\begin{cases}
2^f
&
\begin{aligned}
&\textnormal{if the evaluation is successful on the entire }
\\[-0.2em]
&\textnormal{test set $\mathcal{T}$ within the timeout,}
\end{aligned}
\\[0.4em]
0.0
&
\textnormal{otherwise.}
\end{cases}
\end{equation}
where $f$ is the respective feature, i.e., always avg\_log10\_speedup except for Experiment 3. The choice of \added{\eqref{def:completescore}} is \added{heuristic and} almost purely ad-hoc and goes back to a few \added{trial-and-error} experiments \added{with a small number of iterations only in which we got the impression that this choice is likely advantageous compared to using the metrics directly}. 
\added{In OpenEvolve, each evaluation is assigned a timeout (here: 600 seconds); this means that programs which are not successfully evaluated within this time---e.g., because they freeze due to an error or simply take too long---are terminated and treated as faulty, i.e., assigned combined\_score=$0.0$; cf. \eqref{def:completescore}. This mechanism does not only make the overall process more robust against faulty programmes, but also, in particular, rules out theoretically correct but impractically expensive candidates---in our case, e.g., finding a FLOP-optimal contraction sequence by brute-forcing all possible permutations.}

\subsubsection*{Influence of different LLMs (Experiment 1)}

In this first experiment we compare the capabilities of different open-weight LLMs on behalf of the set of small tensor networks and evolution w.r.t. avg\_log10\_speedup. It can be seen that the models of OpenAI performed best, being the only ones that reach an improvment over the baseline, i.e., avg\_log10\_speedup $> 0$ with GPT-OSS-20B with ``high'' reasoning effort. Surprisingly the smaller 20B-model was able to outperform the larger 120B-model, and even much larger competing models. Among the remaining models, only Qwen3.5-122B-A10B, MiniMax M2.5/M2.7, and the Gemma-4 models, i.e., considerably newer and/or larger models, were able to reach comparable final scores. Some models, e.g., Qwen3 235B or Mistral Large 3, were not able to generate a (correct) program that improved significantly beyond the initial program (avg\_log10\_speedup$\approx -1.31$). As GPT-OSS-20B performed best in this initial experiment and \added{is} extremely cheap to host, we perform the following more extensive experiments with this model only. Without wanting to judge the performance of individual models specifically, we can summarise this experiment as follows: In this single-run experiment, the choice of the model had a large influence on the quality of the result.

\begin{figure}[t]
    \centering
    \includegraphics[width=\columnwidth]{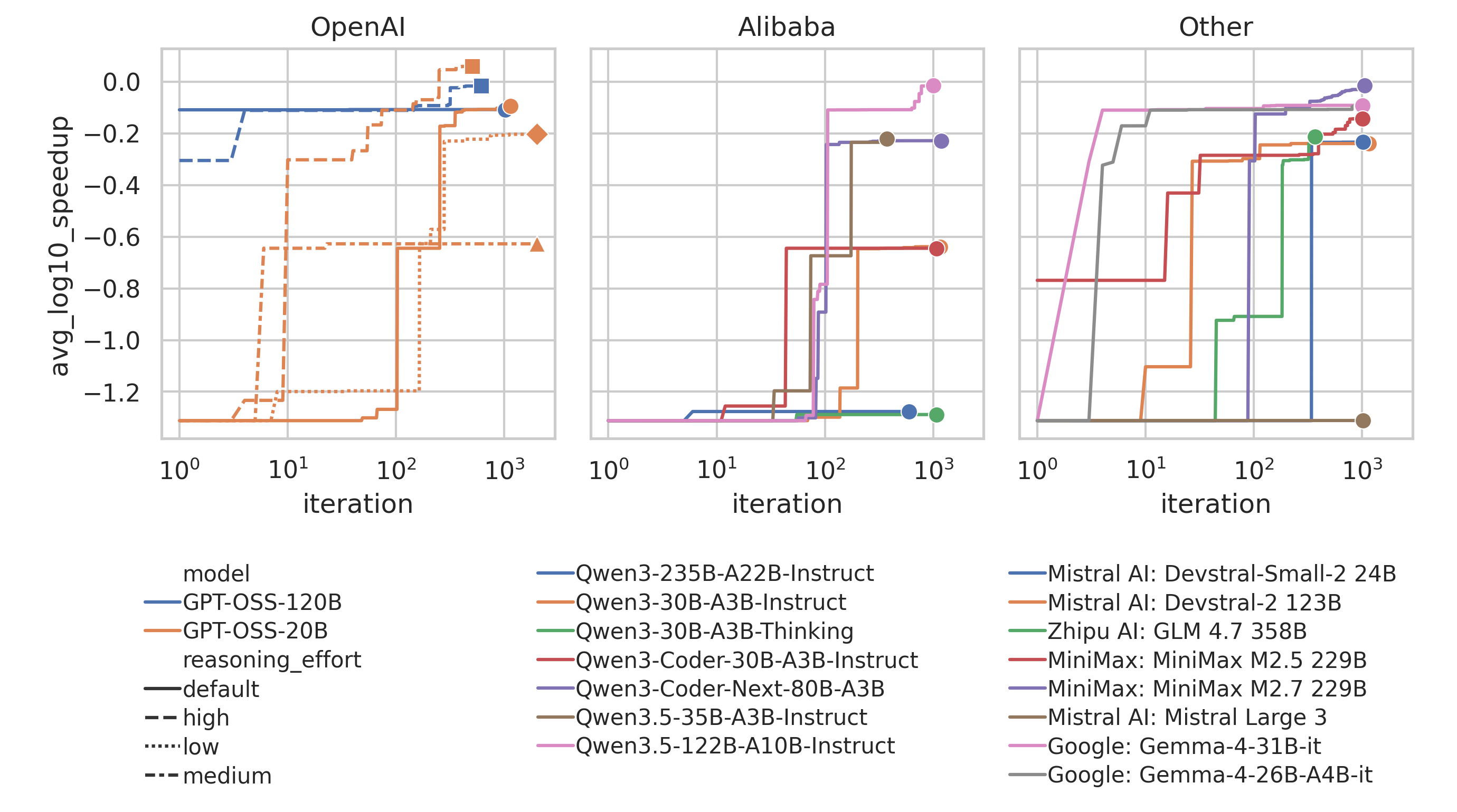}  
    \caption{\added{Experiment 1 ---} Evolution of the average reduction of log$_{10}$ FLOPs for various LLMs on the ``full small'' TNs set.}  
    \label{fig:all_models}
\end{figure}

\subsubsection*{(Non-)Reproducibility and Stochasticity (Experiment 2)}

\begin{figure}[t]
    \centering
    \includegraphics[width=0.8\columnwidth]{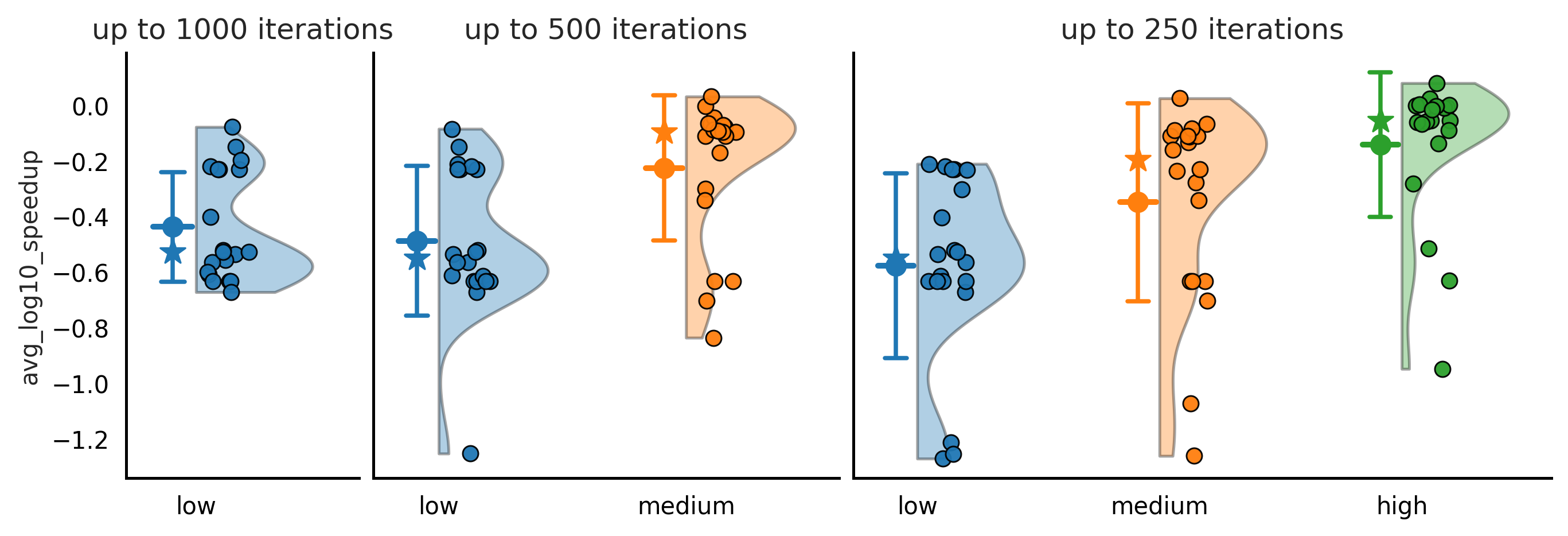} 
    \caption{\added{Experiment 2 ---} Distribution of average reduction of log$_{10}$ FLOPs on the ``small'' TNs over 20 runs with different random seeds after 1000, 500, and 250 iterations, respectively. The LLM is GPT-OSS-20B with different levels of reasoning effort (``low'' to ``high''). The bar left to the violin plot indicates mean (circle), standard deviation, and median (star).} 
    \label{fig:reproducibility_violins}
\end{figure}

Reproducibility is an obvious issue when dealing with LLMs as their text generation is inherently stochastic. To assess whether the outcome of the evolutionary procedure is at least reproducible in a statistical sense, we run the same 3 experiments \added{(``small'' TNs and GPT-OSS-20B with ``low'', ``medium'' and ``high'' reasoning effort)} 20 times \added{each}. The distributions of best achieved metrics in Figure \ref{fig:reproducibility_violins} seem to become tighter for an increasing number of iterations, and a trend of higher scores for higher reasoning effort seems to be on the horizon. Notably, the improvement over the baseline observed in Experiment 1 could be reproduced in some cases. We are tempted to suggest the following interpretation of our observations: Individual evolution runs tend to be highly stochastic, while certain commonalities seem to emerge across an ensemble of different runs.

\subsubsection*{Influence of different Metrics (Experiment 3) and Test Cases (Experiment 4)}

\begin{table}[t]
\small \centering
\begin{tabular}{lccccc}  \hline 
 & \multicolumn{5}{c}{\bf metric measured} \\ \cline{2-6} 
\bf metric optimized \hspace{0.25cm} & min & avg & median & max & total \\ \hline 
 min &  -0.17 &  +0.05  &  +0.05 & +0.36  &  +0.07  \\
 avg & -0.44 &  +0.06 &  +0.06 &  +0.42 & +0.07 \\
 median & -0.76 &  -0.01 &  -0.01&  +0.27&  -0.01 \\
 max & -1.06& -0.16& -0.12&   +0.40& -0.24  \\
 total & -0.47& +0.00 & +0.01 & +0.32 & +0.02  \\ \hline 
\end{tabular}\normalsize
\caption{\added{Experiment 3 --- Final measured metrics (``min'', ``max'', ``avg'', ``median'' as introduced below \eqref{def:log10speedup} and ``total'' as defined in \eqref{def:log10reduction}); larger is better.}}
\label{tab:metrics}
\end{table}

\begin{table}[t]
\small \centering
\begin{tabular}{lccc}  \hline 
& \multicolumn{3}{c}{\bf optimization on} \\ \cline{2-4} 
\bf evaluation on \hspace{0.25cm} & small & middle & large \\ \hline 
small & +0.05 ($\pm 0.09$) & -0.02 ($\pm 0.09$) & -0.24 ($\pm 0.13$) \\
middle & +1.05 ($\pm 1.21$) &  +1.61 ($\pm 1.07$) & +0.20 ($\pm 0.89$) \\
large & $\star$ & $\star$ &  -0.23 ($\pm 0.05$) \\ \hline
\end{tabular}
\caption{\added{Experiment 4 --- Final average reduction of log$_{10}$ FLOPs (larger is better, $\pm$ indicates standard deviations); ``$\star$'' indicates that the evaluation did not terminate in time.}}
\label{tab:sets}
\end{table}

In order to assess the influence of the metric chosen to define the combined score, we run experiments on the set of ``small'' TNs for each possible choice while tracking all available metrics. The results \added{of Experiment 3} are \added{summarized} in Table \ref{tab:metrics}. As key observation we may regard that even when evaluated on a fixed set of test examples and without measurement errors, metrics can differ in terms of how well they can be optimized by OpenEvolve, and to which extent their optimization simultaneously results in an improvement of other, also potentially useful, metrics. This should in particularly be seen in the light of the fact that a number of rather strong conceptual assumptions have already been made before with regard to the definition of the metrics, e.g., the restriction to pure FLOP counts.  
\added{In Experiment 4,} we fix the metric to be optimized again to ``avg\_log10\_speedup'' and investigate the influence of the set of test examples \added{by optimizing} over ``small'', ``middle'' and ``large'' TNs, respectively, while evaluating also on the \added{other} sets. As this setup is similar to the split of training and test/validation data in deep learning \cite{Goodfellow-et-al-2016} one may expect typical in-distribution vs out-of-distribution phenomena \cite{yang2024generalized}; \added{our experimental results summarized in Table \ref{tab:sets} appear to roughly support this hypothesis. In any way,} the selection of test examples for the evaluation of the LLM-generated programs seems to strongly influence the outcome of evolution. 
Altogether, Experiments 3 and 4 suggest that the following two questions are crucial for success of the evolutionary algorithm: How to define and measure ``improvement''?---in other words: Which metric to optimize and how to evaluate this metric, i.e., on behalf of which examples etc.?

\subsubsection*{Detailed Examination of the ``best'' Outcome}\label{subsec::best}

\begin{figure}[h!]
    \centering
    \includegraphics[width=0.77\columnwidth]{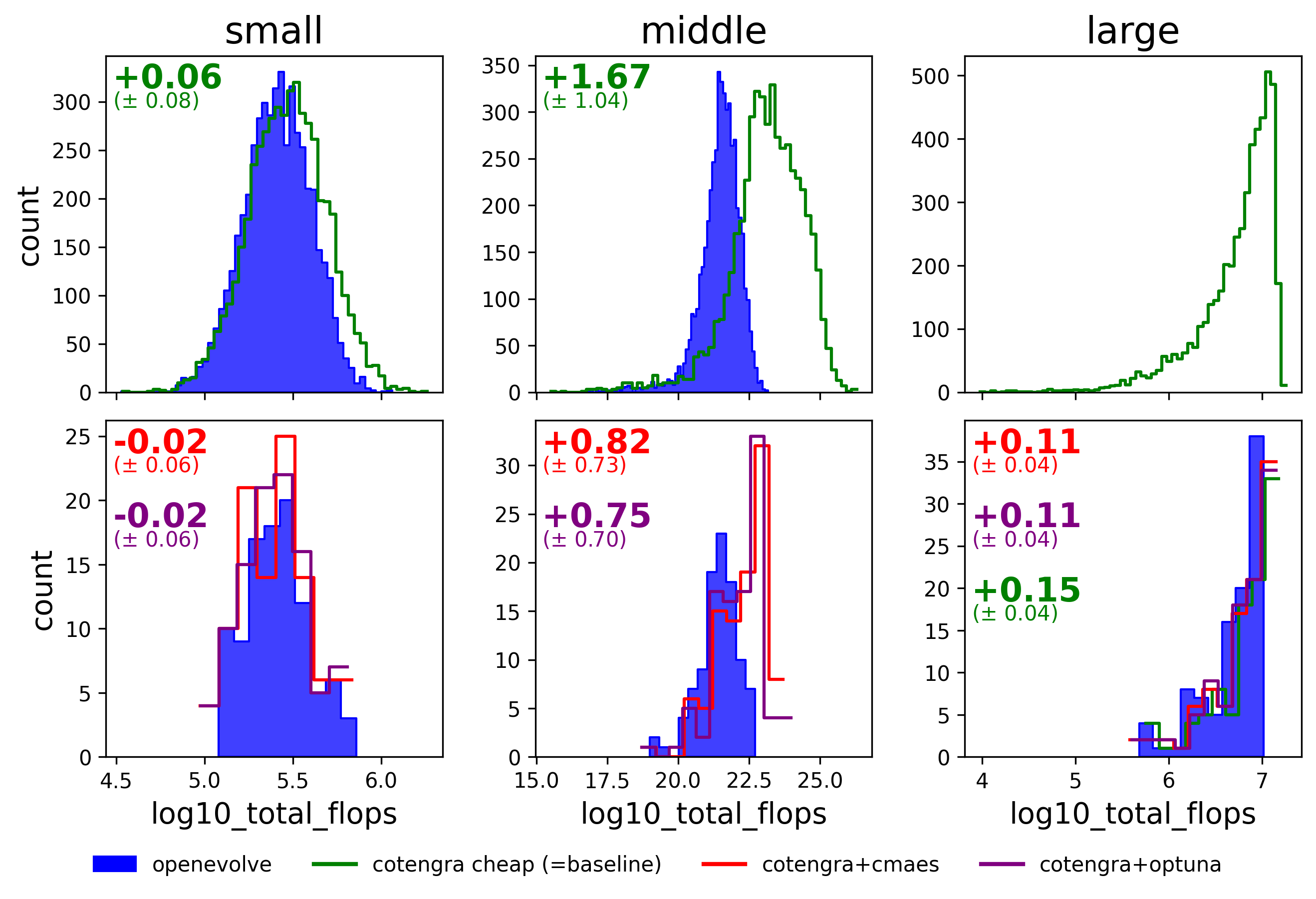} \includegraphics[width=0.77\columnwidth]{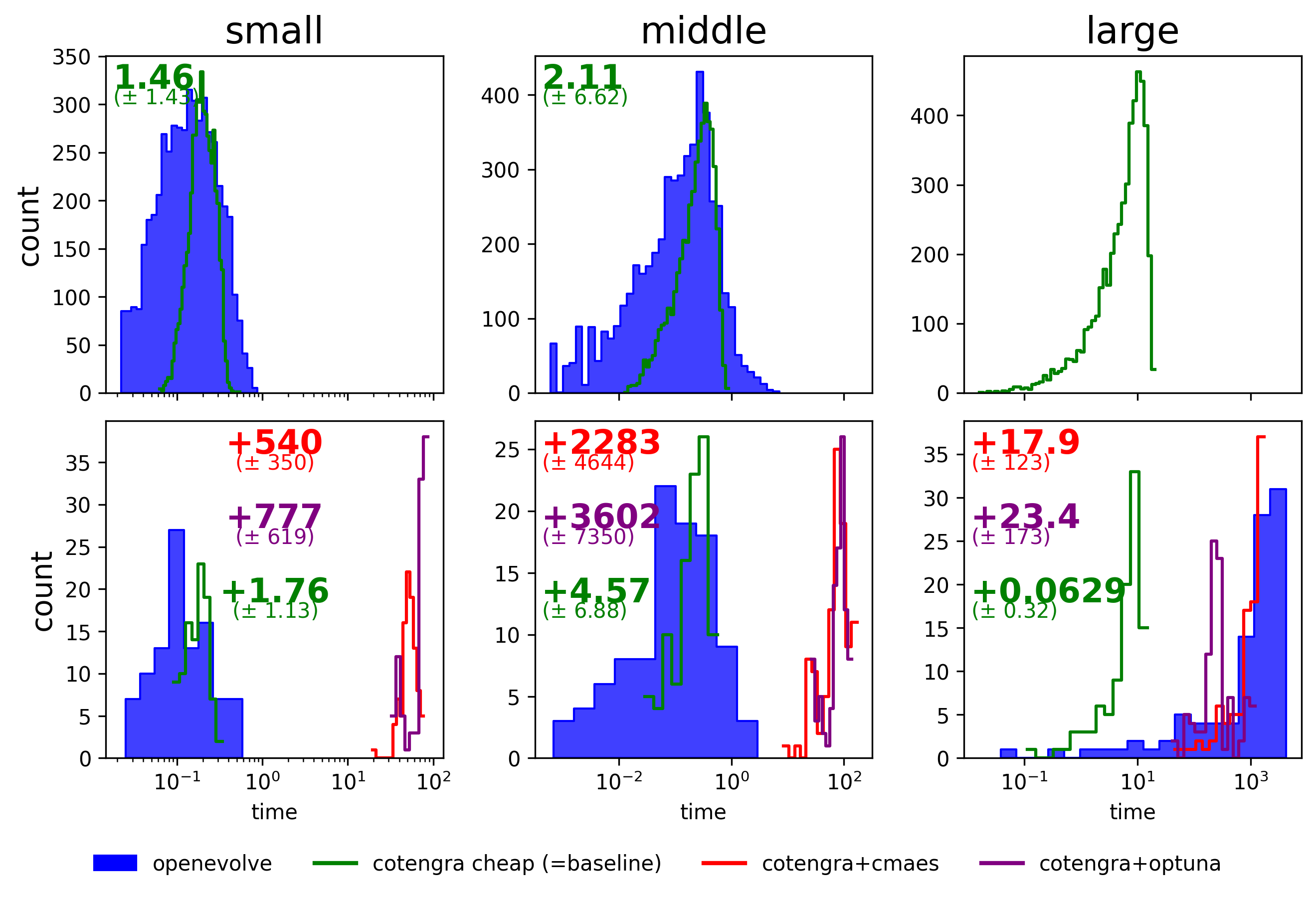}   
    \caption{\added{Experiment 1 ---} Distribution of log$_{10}$ FLOPs (top) and  distribution of runtimes for computing the contraction order (bottom) for the final code of the best run (blue) vs different baselines (``cotengra cheap'', ``cotengra+cmaes'', and ``cotengra+optuna'' in green, red, and purple, respectively) on different TNs sets (columns). First and third rows use the full TNs sets, second and fourth row use only the last 100 of the TNs from the respective sets. The colored numbers give the average improvement of $\log_{10}$ FLOPs (higher is better, $>0$ means improvement) or the average reduction factor of runtime (higher is better, $>1$ means improvement) vs the respective baseline on the respective set of TNs; \added{``$\pm$" indicates standard deviations.}} 
    \label{fig:actual_speedup}
\end{figure}

Finally, we examine the best outcome of Experiment 1 (by GPT-OSS-20B, reasoning level ``high'') in more detail. In order to further assess the quality of the final code, we compare it to stronger baselines and also take the run time of the contraction order determination itself into account, both on ``small'', ``middle'' and ``large'' TNs. The fact that already the run time of the OpenEvolve-generated algorithm (not of the actual contraction) is only feasible for ``small'' and ``middle'' TNs \added{(cf. also Experiment 4)} underpins the relevance of this consideration for practical purposes. 

Figure \ref{fig:actual_speedup} shows that our final program \added{clearly} outperforms the baseline ``cotengra cheap'' on ``middle'' TNs (average reduction of FLOPs by a factor of $\sim 10^{1.67} \approx 47$) while the effect is smaller, but still positive on ``small'' TNs (average reduction of FLOPs by a factor of $\sim 10^{0.06} \approx 1.15$). For ``large'' TNs, however, we could not obtain results as the run time of the algorithm in the final program was too high. This highlights another important fact: the combination of test examples (in our case: ``small'' TNs) and the timeout specified for the evaluation also formulates an implicit constraint on the run time of the program to be optimized. Obviously, choosing ``large'' TNs as test examples would have resulted in a different final program as the present final program would not have been evaluated successfully in that setting. In order to take into account the \added{much} higher computing time on ``large'' TNs and more expensive baselines, we only perform the corresponding analysis on 100 TNs of the respective TN quantities; see the second row in Figure \ref{fig:actual_speedup}. Compared to the stronger baselines ``cotengra+cmaes'' and ``cotengra+optuna'', there is still a \added{clear} improvement on ``middle'' TNs (average reduction of FLOPs by factors of $\approx 6.6$ and $\approx 5.6$), while we cannot clearly interpret the results on ``small'' and ``large'' TNs. In terms of run times (Figure \ref{fig:actual_speedup}, bottom), the program generated by OpenEvolve is slightly faster than contraction order determination by the baseline ``cotengra cheap'', while on ``large'' TNs the situation is completely opposite and our program exhibits run times exceeding reasonable limits for experimentation. On the subsets of 100 TNs, respectively, it turns out that the generated algorithm is clearly faster than the stronger baselines. However, our experiments do not take into account the potential of  parallelization for these search-based baselines that might heavily reduce the observed gap. On ``large'' TNs, finally, the adverse run time behaviour of the algorithm becomes apparent when it is clearly outperformed (in terms of run time) by the baselines while having a comparable quality of the outcome (in terms of FLOPs of the resulting paths, cf. Figure \ref{fig:actual_speedup} \added{(top)}). \added{Finally, we recall that the actual speed-up in real-world applications will not depend solely on the number of FLOPs. Moreover, both the resource utilization of the contraction itself and the determination of the contraction order must be taken into account to determine a true benefit for a particular use-case. Both aspects go beyond the scope of this paper. Nevertheless, our results indicate in our view} that careful (experimental) validation is just as important as the well-considered design of the harness for evolution.

In contrast to, e.g., the training of a neural network, the result of evolution in OpenEvolve is typically not a black-box, but a human-readable code, \added{so} it can be subject to a theoretical validation (in addition to an experimental, practical validation) as well. In the following, we therefore examine the overall best code in detail. \added{Before doing so, let us \added{briefly} mention in this context that autonomously AI-generated code---and, as we believe, algorithms, too---can be difficult to understand as it might exhibit properties known for making code comprehension challenging: It is not self-written and the context might not be fully known; thus, building a mental model is more difficult and understanding might have to happen bottom-up instead of top-down; cf., e.g., \cite{Feitelson2022,Siegmund2015,Agrawal2024,Goncalves2025,Neumueller2025}. In the context of research software, this is a particularly serious issue, as comprehensibility/understandability should not be a ``nice to have'' but an important quality requirement; cf. the emphasis on understandability, documentation, reuse, evolvability, and adaptability as demands for sustainable research software in \cite{Felderer2025}. Nevertheless, it should also be noted that these difficulties naturally go hand in hand with the greatest (at least hoped-for) advantage of AI-based algorithm engineering: the users are hoping for ideas or approaches, e.g., from other specialist areas, that are not yet known to them or the discovery of genuinely new procedures. For an analysis of code complexity of the codes generated in Experiments 1 and 2, we refer the interested reader to Appendix \ref{subsec::complexity}.}

OpenEvolve did not change the docstring of {\tt find\_edge\_path} but turned the original 4 lines of code (including 1 line of comments, doc string excluded) into 164 lines of code (including 21 lines of comments, doc string excluded). Thus, the comments are rather thin on the ground (cf. also Appendix \ref{subsec::complexity}), and, in addition, they do not contain any references to the literature at all. Consequently, an assessment of novelty of the algorithmic core had to be done from scratch. 

It turns out that the overall skeleton of \texttt{find\_edge\_path} is well known.  In particular, tensor-network contraction planning can be formulated as a \emph{graph elimination} (variable-elimination) problem: eliminating a vertex induces \emph{fill-in} edges among its neighbours, and the resulting induced graph controls intermediate tensor growth and cost. In the tensor-network setting, optimal contraction complexity has been related to the treewidth of the network's \emph{line graph}, a viewpoint that can be expressed directly in terms of elimination orderings and fill-in \cite{MarkovShi2008,Dumitrescu2018TreewidthBenchmark}.
The routine follows this paradigm by building an \emph{index interaction graph} whose vertices are index labels and where two indices are adjacent if and only if they co-occur in a tensor (equivalently, each tensor induces a clique on its incident indices); this is essentially the line-graph/interaction-graph representation well-known from treewidth-based formulations \cite{MarkovShi2008,Dumitrescu2018TreewidthBenchmark}. It then computes an elimination ordering using a greedy \emph{min-fill}-style rule and explicit fill-in updates \cite{rose1976vertex}, followed by a lightweight local refinement that repeatedly swaps adjacent positions in the ordering (a simple local-search strategy for improving triangulations/orderings) \cite{kjaerulff1990triangulation}.  In contrast, many modern tensor-network planners, e.g. \cite{GrayKourtis2021hyperoptimized}, optimize in the space of contraction trees/hypergraph contractions using randomized and local reconfiguration strategies rather than operating purely on an elimination ordering.

The nonstandard aspect lies in the \emph{particular surrogate weighting} of prospective fill edges: these are penalized by $\min(\log \dim a,\log \dim b)$ (with $\dim(\cdot)$ the \emph{index} dimensions), whereas standard \emph{weighted} min-fill variants typically weight fill edges by multiplicative measures of variable cardinalities (equivalently, additive in log-space) \cite{kjaerulff1990triangulation,dechter1999bucket}.  We were not able to find prior tensor-network contraction heuristics that employ this ``minimum-of-logs'' fill weighting specifically, and, additionally, we are not aware of a theoretical justification of this. Consequently, if the results shown in Figure \ref{fig:actual_speedup} would not have indicated that the algorithm indeed yields reasonable results, one could regard this choice as an error.

\section{Conclusion}\label{sec::conclusion}

During this case study we were not able to discover a significantly new algorithm. Instead, we obtained known approaches with minor modifications. However, the clear limitations of this work should be taken into account: we performed our experiments with a single framework for LLM-based algorithm evolution only---and, in fact, one that is not a very autonomous agentic system and that exclusively relies on the internal knowledge of the LLM. In addition, we restricted ourselves to a single, very specific and rather easy to handle, though interesting, example task, too. Moreover, we did not make use of commercial, closed-source frontier LLMs (which could potentially have delivered better results), and the number of experiments was limited to a feasible\footnote{\added{``Feasible'' refers particularly to a {\itshape fair} usage of the {\itshape shared} LLM-providers Blablador and ChatAI.}} amount of compute budget. Finally, both LLM capabilities and surrounding frameworks are evolving extremely fast; so one might consider the setup of the shown experiments as outdated quite soon.     

Nevertheless, we believe that some higher-level lessons can be learned. We regard OpenEvolve and similar or more advanced systems as an exciting, highly promising approach to scaling algorithm search and optimization through (more or less) brute-force trial-and-verify-guided search. A major advantage---and at the same time a major challenge in subsequently understanding the generated code---is the enormous cross-domain knowledge base of LLMs. The design of the space within which the LLM is given freedom of design, so to speak, remains crucial: the human scientist must condense the overarching algorithm optimization goal into a suitable evaluation function; even in a comparatively simple case such as the example discussed here, there are numerous design decisions that can have a significant influence on the final result, as we have exemplarily seen in \added{Experiments 3 and 4}. It should be noted that building such an evaluation function might even be more complex for different problem types where evaluation might require compilation and/or actual run time- and/or memory measurements, possibly on certain hardware configurations. In addition, experimental validation of the outcome of the evolution process, beyond those characteristics covered by the bespoke evaluation function, i.e. testing for generalization, as well as interpretation and theoretical understanding are still necessary and can be non-trivial. This might become even more important in the context of advanced agentic frameworks with a less rigid structure than OpenEvolve which could be susceptible to reward-hacking-type phenomena; cf., e.g., \cite{wen2026automatedw2s}. We believe that at least these two tasks, designing the harness as well as carrying out final validation and interpretation, both taking into account the overarching goal(s) and motivation, remain in the human scientists/(research) software engineers responsibility for the foreseeable future. 

With regard to the consumption of resources, it should be noted that in addition to the not inconsiderable (energy and financial) costs of LLM use \cite{caravaca2025promptspowermeasuringenergy}, the costs for the evaluation must also be taken into account: here there is a tension between the most comprehensive and good evaluation possible and the higher duration or other resource utilization of such an evaluation that may be necessary for this. Another aspect that deserves attention is the choice of the underlying LLM: at least in our experiments, this strongly influenced the result. This opens up a further area of tension, in that LLMs with stronger capabilities are often, though not necessarily, larger and more expensive to operate, or even only available commercially and operated abroad, which ultimately also potentially touches on the extremely sensitive issues of technological sovereignty, export control and data security. Analogous concerns regarding the threat of a ``digital divide'' between ``AI-have'' and ``AI-have-nots'' in the mathematical sciences were also expressed in \cite{klowden2026mathematicalmethodshumanthought}.

To conclude the paper, let us try to put our previous comments in a slightly broader context. Trying to view our setup through the conceptual lens of \cite{feldt2026semiexecutablestackagenticsoftware}, the final code for tensor network contraction is only the innermost executable artifact; system message, OpenEvolve workflow, evaluation metric, the set of test examples, timeout configuration, as well as post-hoc validation and interpretation can be seen as elements of the ``control systems'' ring, as ``semi-executable'' specifications of the scientific objective(s). In this sense, our experiments with a subset of these specifications can be understood as illustration of the claim made in \cite{feldt2026semiexecutablestackagenticsoftware} for agentic software engineering in general, now of course rephrased to our scientific algorithm-/performance-engineering setting: prompts, agentic workflows/frameworks, evaluation-/scoring-related artifacts and routines, as well as the setup, the results, and their interpretation, of post-hoc validation (experimental or theoretical) become first-class engineering objects.

With regard to the concepts formulated in \cite{baeuerle2026}, we are tempted to categorise our results as follows: the set screws examined in numerous experiments illustrate the central importance of ``intentmaking'', but also the associated challenges. In our case, we would understand ``intentmaking'' as the process of condensing an initially open, not precisely formulated algorithm improvement task into an input suitable for OpenEvolve, which definitely requires a set of non-trivial design decisions. We also tackled the step ``sensemaking'', at least partially, in the form of a more detailed post-hoc analysis of the best generated candidate. In order to achieve a result that is actually relevant for domain science or application, the loop would have to be closed and further steps of ``intentmaking'' (e.g. by selecting new test cases, including the runtime of the ordering algorithm in the score function, choosing a performance metric beyond pure FLOP count, etc.) and ``sensemaking'' would have to follow\footnote{In fact, a tiny and hidden approach to this has already happened when we chose $2^f$ in the definition of combined\_score based on a few experiments as indicated at the beginning of Sect. \ref{sec::experiments}}. We also agree with \cite{baeuerle2026} that working with OpenEvolve on the problem at hand is better understood as experimenting with, and simultaneously designing, a complex scientific instrument than as using a simple question-answer oracle.


\begin{acknowledge}
The usage of Blablador API, provided by Helmholtz AI Jülich via HIFIS, and SAIA/ChatAI API, provided by GWDG Göttingen via Academic Cloud, is gratefully acknowledged. Moreover, the authors gratefully acknowledge the scientific support and HPC resources provided by the German Aerospace Center (DLR). The HPC system CARO is partially funded by ``Ministry of Science and Culture of Lower Saxony'' and ``Federal Ministry for Economic Affairs and Climate Action''. This project was made possible by the DLR Quantum Computing Initiative and the Federal Ministry for Economic Affairs and Climate Action; \url{https://qci.dlr.de/qutenet/}. \added{Finally, the authors thank the anonymous reviewers for their feedback that helped to improve the quality of the paper.}
\end{acknowledge}
\begin{data}
Scripts used for the generation of tensor networks, an example set up configuration for the OpenEvolve experiments, and the final codes of Experiment 1 are available on Zenodo: \url{ https://doi.org/10.5281/zenodo.22108774}
\end{data}
\begin{aiuse}
Besides its obvious usage in the experiments, generative AI has been used for coding, formatting, drafting, and assistance during literature research. AI-generated content created in the course of this was checked, corrected if necessary, and revised by the authors.
\end{aiuse}

\bibliographystyle{eceasst}
\bibliography{bibliography.bib}

\newpage
\appendix 
\section{Appendix}

\subsection{Additional Details on the Setup of the Experiments}
\label{subsec::setup}

Cotengra's function {\tt utils.lattice\_equation} is used to generate random tensor networks of type ``scalar product of two periodic MPS'' with lengths and dimensions within specified bounds to obtain the sets of ``small'', ``middle'', and ``large'' tensor networks, each of them containing 5000 networks; for usage in the different experiments they are saved to {\tt json} files. \added{To reduce computational costs of the evaluation function,  we sometimes use smaller sets of TNs in which we only use the first 1000 TNs from the aforementioned sets and refer to them as ``reduced'' sets, whereas the original ones are referred to as ``full'' for clarity. For Experiment 4 we also consider a mixed set named ``all''. A summary of all sets is provided in  Table \ref{tab:datasets}. } The reference contraction orders are computed by cotengra's HyperOptimizer; see Table \ref{tab:baselines} for the respective configurations. \added{Hereby, no parallelization is utilized.} 

The configuration of OpenEvolve for all experiments is shown in Table \ref{tab:config}; all parameters not mentioned in this table are set to their default values\footnote{\href{https://github.com/algorithmicsuperintelligence/openevolve/blob/main/configs/default_config.yaml}{https://github.com/algorithmicsuperintelligence/openevolve/blob/main/configs/default\_config.yaml } [accessed 04/2026]} (e.g., temperature$=0.7$ and top\_p$=0.95$ for the LLM). The system message and the initial program are found as Listings \ref{systemmessage} and \ref{initialprogram}, respectively; the latter implements a  \added{highly naive} greedy approach in which indices are contracted in the order of their size (descending). An overview over the open-weight LLMs used in the experiments and how they were accessed can be found in Table \ref{tab:models}. 

Experiment 1 as well as the analysis of the respective outcomes were performed on Linux Workstations under Python 3.10 whereas Experiments 2, 3 and 4 were conducted on nodes of DLRs cluster CARO equipped with a single Nvidia GH200 under Python 3.12.

\subsection{Additional Details on Experiment 1}

If the evolution trajectories \added{shown in Figure \ref{fig:all_models}} for certain models end before 1000 iterations, this is due to excessive time consumption and/or a high number of restarts from checkpoints\footnote{\added{Restarts become necessary if errors/timeouts during evaluation are not catched successfully by the timeout mechanism in OpenEvolve resulting in crashing/freezing of the entire evolution process.}} due to frequently crashing/timing out evaluations \added{in which case the experiments were aborted manually}; the latter issue happened, e.g., for the GLM 4.7 model. \added{While referring to iteration numbers is common and convenient, it should be noted that resource usage per iteration varies considerably between the shown runs: the time for evaluation ranges from a few seconds (for the initial code) to a maximum of 10 minutes in case of a timeout. On the LLM side, the number of output tokens, the hardware required and the time needed for generation depend on the model, the deployment and the system load. The configuration in OpenEvolve, however, at least yields an upper bound of resource utilization per iteration in terms of time and token usage; cf. Table \ref{tab:config}.}

\subsection{Additional Details on Experiment 2}

\begin{figure}[h]
    \centering
    \includegraphics[width=0.8\columnwidth]{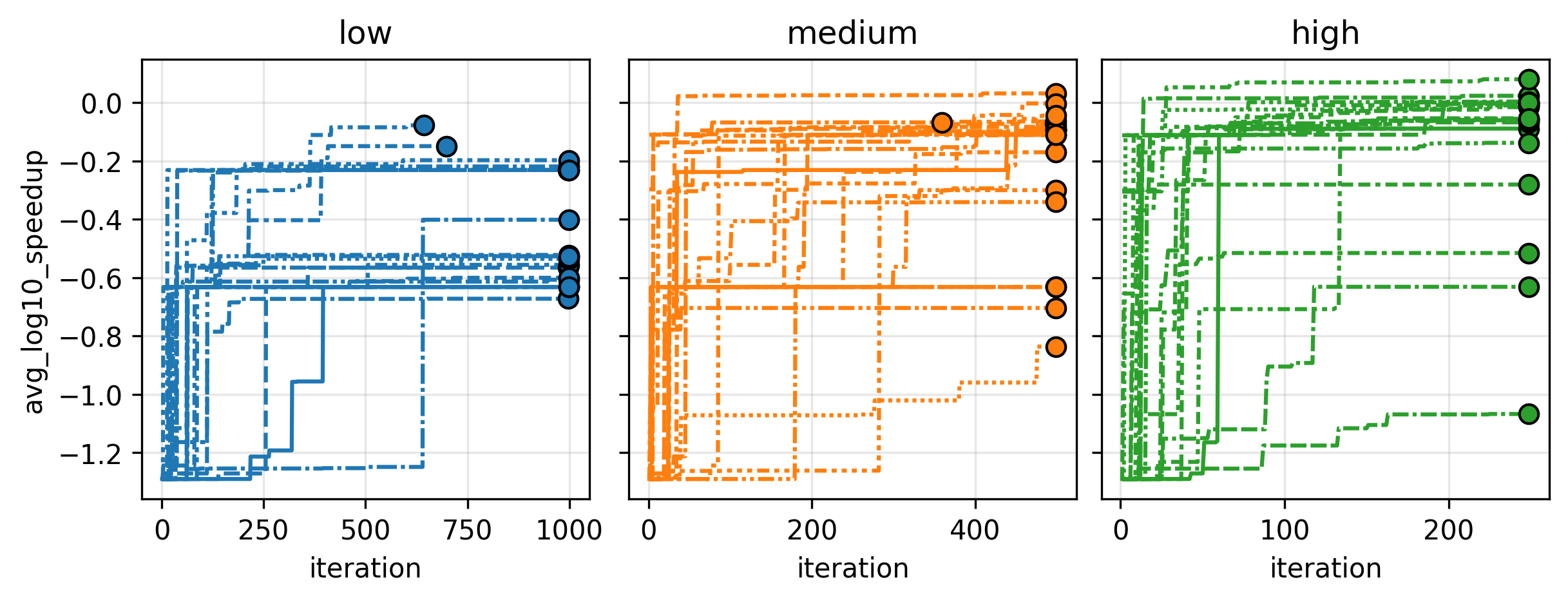}
    \caption{\added{Experiment 2 --- Evolution of the average reduction of log$_{10}$ FLOPs on the ``reduced small'' TNs set over 20 runs with different random seeds. The LLM is GPT-OSS-20B with different levels of reasoning effort (``low'' to ``high'').}} 
    \label{fig:reproducibility_paths}
\end{figure}

OpenEvolve allows setting seeds that are respected both in its evolution algorithm and fed forward into the LLM-API call; it should be noted that this still might not guarantee reproducibility on the LLM-side \cite{atil-etal-2025-non,yuan2025understanding}. We run the same 3 experiments 20 times with different seeds $0$-$19$ (instead of OpenEvolve's default 42). The decision to perform 250, 500, and 1000 iterations, respectively, is due to the fact that the different reasoning efforts ``high'', ``medium'', and ``low'' for GPT-OSS-20B came with a roughly similar ratio in terms of computing time \added{for the whole evolution process}. \added{For a different number of iterations, the explanation is otherwise the same as for Experiment 1 (see above).} Nevertheless, to keep the computational effort reasonable we had to switch from the full set of small tensors to the reduced one (see Table \ref{tab:datasets}) \added{and run evaluations on 4 cores only. Under these changes, the fixed timeout of 600s for each evaluation places slightly different demands on the programmes being evaluated than in the previous experiment, of course. In addition to the distribution of average reduction of $\log_{10}$ FLOPs shown in Figure \ref{fig:reproducibility_violins}, we show here also the evolution trajectories leading there in Figure \ref{fig:reproducibility_paths}.}

\subsection{Additional Details on Experiments 3 and 4}

To reduce the computational effort, all runs in Experiments 3 and 4 utilize the reduced TNs sets and GPT-OSS-20B with ``medium'' reasoning effort. For each configuration under consideration ---different metrics over ``small'' TNs in Experiment 3, ``avg\_log10\_speedup'' over different sets of TNs in Experiment 4---10 runs with different seeds $0$-$9$ have been done utilizing the same hardware configuration as in Experiment 2. However, each run has been assigned a hard {\itshape overall} timeout of 12 hours\footnote{including 20 minutes reserved for starting vllm to host the LLM} and in case of crashing before {\itshape no} restart has been done. The evolution trajectories and the reasons for their termination are shown in Figure \ref{fig:exp3and4}. To reduce stochastic effects, for the actual evaluations in Experiment 3 and 4 only the run with the best final score has been selected for each configuration.  

\subsubsection{Additional Details on Experiment 3}
\begin{figure}[h]
    \centering
    \includegraphics[width=\columnwidth]{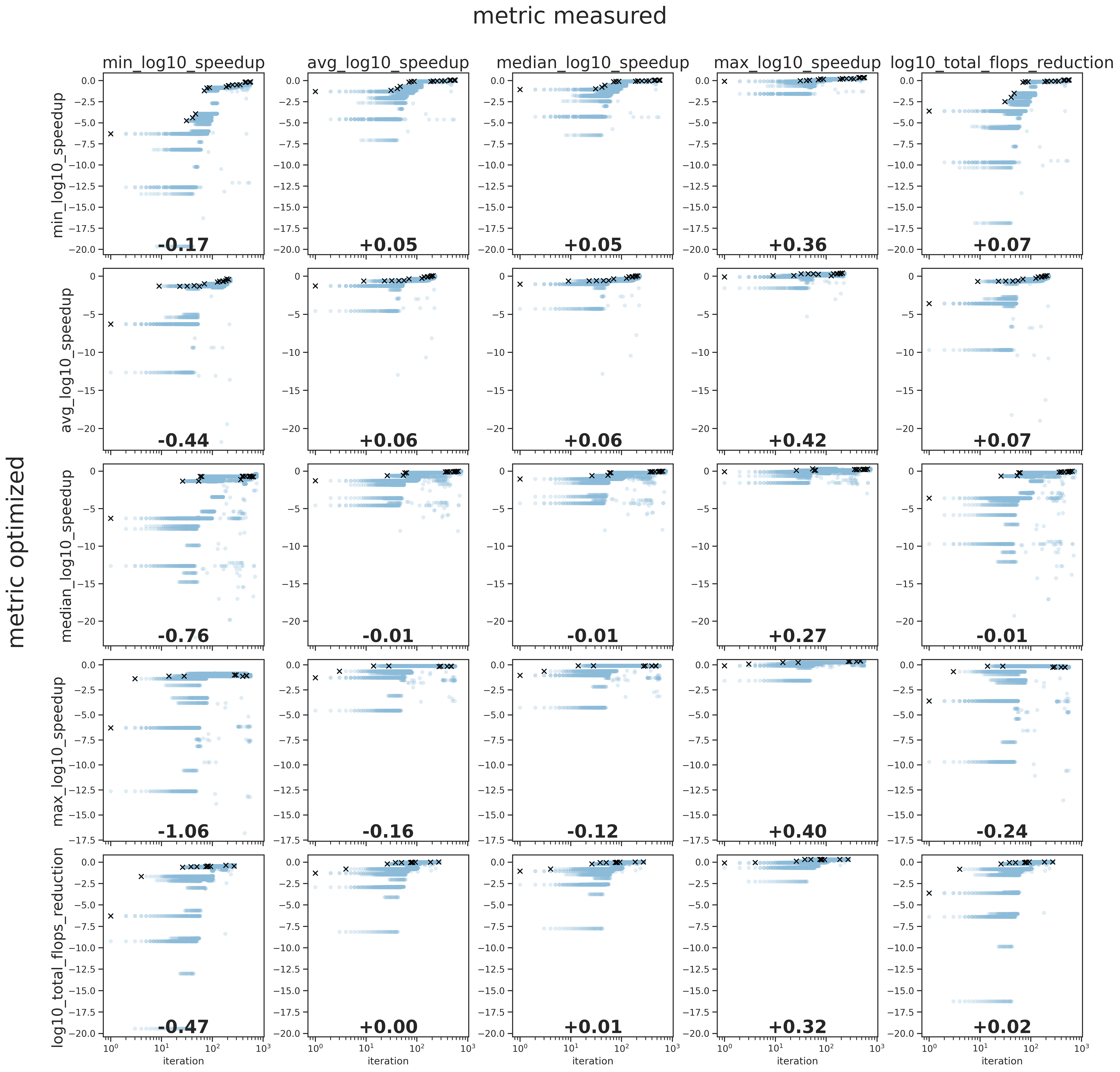}  
    \caption{\added{Experiment 3 ---} Evolution of various metrics (``metrics measured'', columns) over the iterations for five experiments with different metric optimized (``metric optimized'', rows) with GPT-OSS-20B on the reduced ``small'' TNs set. Light blue dots indicate individual members of the population, whereas black crosses indicated members of the population that just have become a new best solution (w.r.t. the metric optimized). The number in the lower part of each plot gives the value of the measured metric for the final best solution for the respective experiment.} 
    \label{fig:metric_vs_metric}
\end{figure}

For this experiment, the description of ``combined\_score'' in the system message (cf. Listing \ref{systemmessage}, line 15) has been adapted accordingly for each chosen metric, of course.  In addition to Table~\ref{tab:metrics} showing the final values of the metrics, Figure \ref{fig:metric_vs_metric} displays the whole evolution process. In most cases, optimization based on a particular metric yields comparatively good results for that metric (first or second place); in addition, optimization against minimum and average seem to perform rather well, whereas optimization against maximum and  median show weaker results. While some observations---such as the rather poor performance of optimization over maximum on the minimum, the mean and the median, are expected and understandable, some others are less: e.g., the comparatively solid performance of ``log10\_total\_flops\_reduction'' can be viewed as surprising, as it is not obvious that this metric generates a reasonable signal for optimization because different contributions of FLOPs from different TNs in the test set might be averaged out. Similarly surprising is the relatively poor performance of optimizing the median compared to optimizing the mean; one could in principle suspect a stochastic effect here, an explanation which, however, is not necessarily supported by the comparatively low variation in the corresponding results shown in Figure \ref{fig:exp3and4}.

\subsubsection{Additional Details on Experiment 4}

\begin{figure}[h]
    \centering
    \includegraphics[width=0.75\columnwidth]{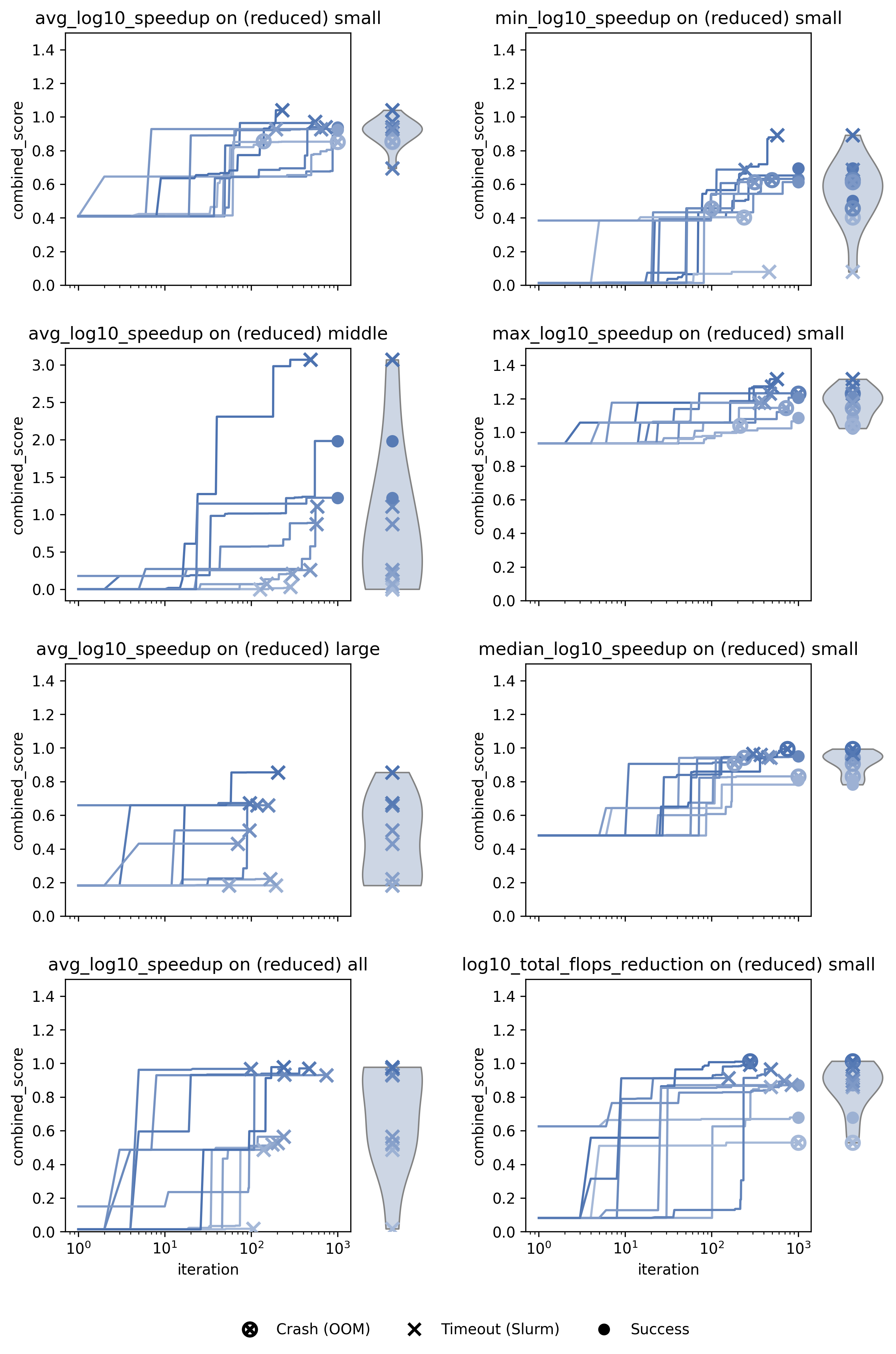}  
    \caption{\added{Experiments 3 and 4 --- Evolution of ``combined\_score'' for 5 runs of the configurations in Experiments 3 (rhs column) and 4 (lhs column). The different markers at the end of the trajectories indicate different reasons for the end of the iteration.}} 
    \label{fig:exp3and4}
\end{figure}

\begin{figure}[h]
    \centering
    \includegraphics[width=0.9\columnwidth]{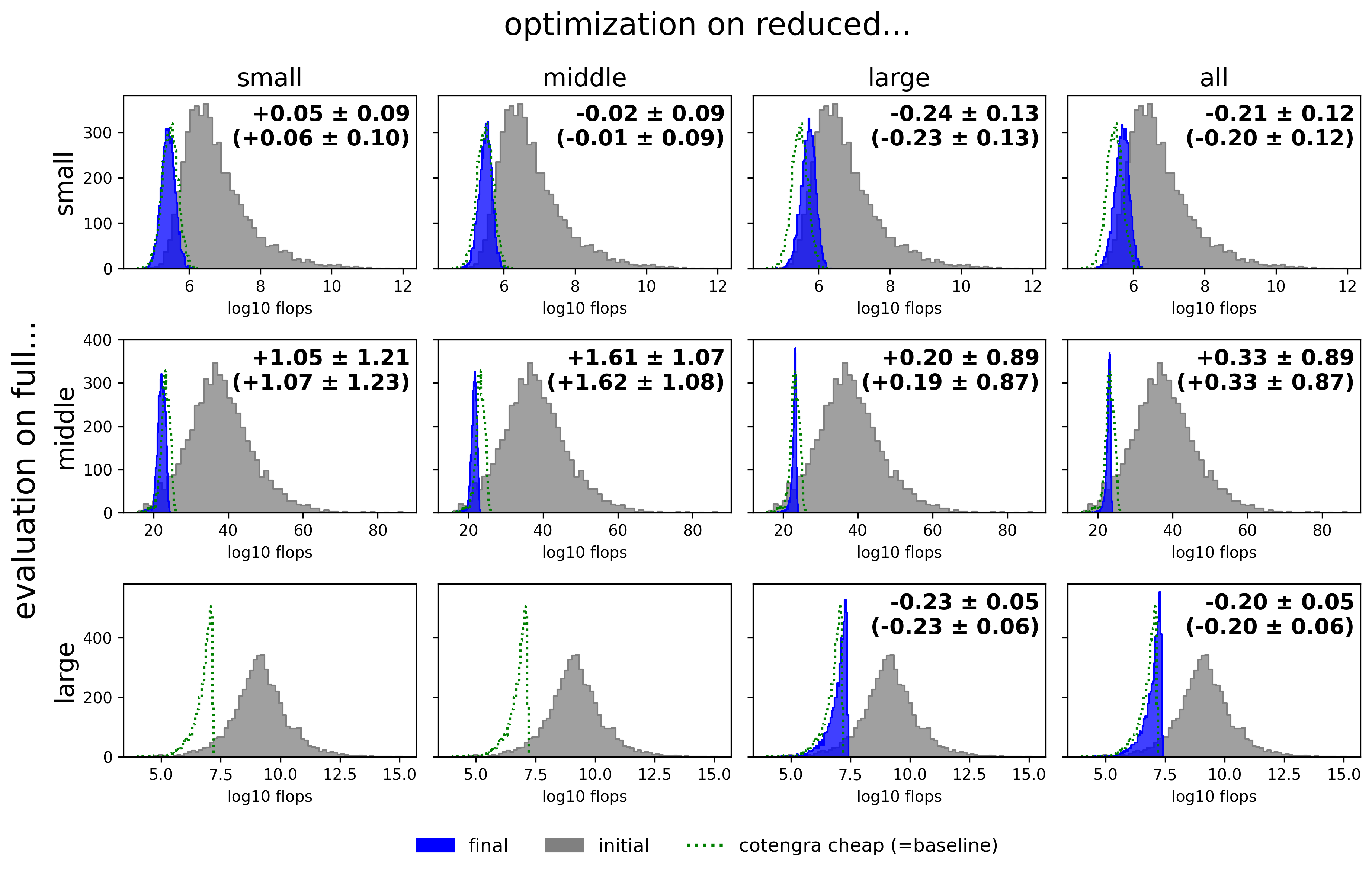}  
    \caption{\added{Experiment 4 ---} Distribution of log$_{10}$ FLOPs for the final best solution of four experiments with GPT-OSS-20B (evolution on reduced ``small'', ``middle'', ``large'' and ``all'' TNs sets; columns), evaluated on the full ``small'' to ``large'' TNs sets (rows). The gray and blue histograms show the distribution of FLOPs of the initial and the final best code, respectively. The dotted green line indicates the distribution observed for the baseline ``cotengra cheap'' and the \added{numbers in the upper right corner give the average improvements compared to this baseline (higher is better) on the full and the reduced (in brackets) sets; $\pm$ indicates standard deviations.} } 
    \label{fig:data_vs_data}
\end{figure}

In this experiment, optimization takes place over the {\itshape reduced} ``small'', ``middle'', ``large'' TNs set (i.e., 1000 networks each), whereas evaluation uses the respective {\itshape full} sets (i.e., 5000 networks each). One additional configuration not yet mentioned in the main text also uses the mixed set ``all'' (cf. Table \ref{tab:datasets}) for optimization. In general, the results of this experiment should be interpreted with the caveat that the corresponding distributions in Figure \ref{fig:exp3and4} appear to be relatively broad. Figure \ref{fig:data_vs_data} shows histograms of the log$_{10}$ number of FLOPs on the three evaluation sets for the final code of the four different configurations vs the initial code and the baseline ``cotengra cheap''. As expected, the programmes perform best on the set of TNs on which they were optimized. This is most obvious in the case where optimization on ``small'' or ``medium'' TNs resulted in programmes that could not even be executed within a reasonable time on the ``large'' TNs set. Moreover, the scores computed on the reduced and the full sets roughly coincide which can be interpreted in that direction that no ``overfitting-like'' effect took place in the in-distribution regime. The fact that optimization on the ``all'' set yielded a competitive result only on the ``large'' TNs set suggests that the three sets of TNs may exhibit characteristics too different for joint optimization, and that the difficulties associated with ``large'' TNs might be dominant.

\subsection{Additional Details on Code Complexity and Comprehension}
\label{subsec::complexity}

\begin{figure}[h]
    \centering
    \includegraphics[width=\columnwidth]{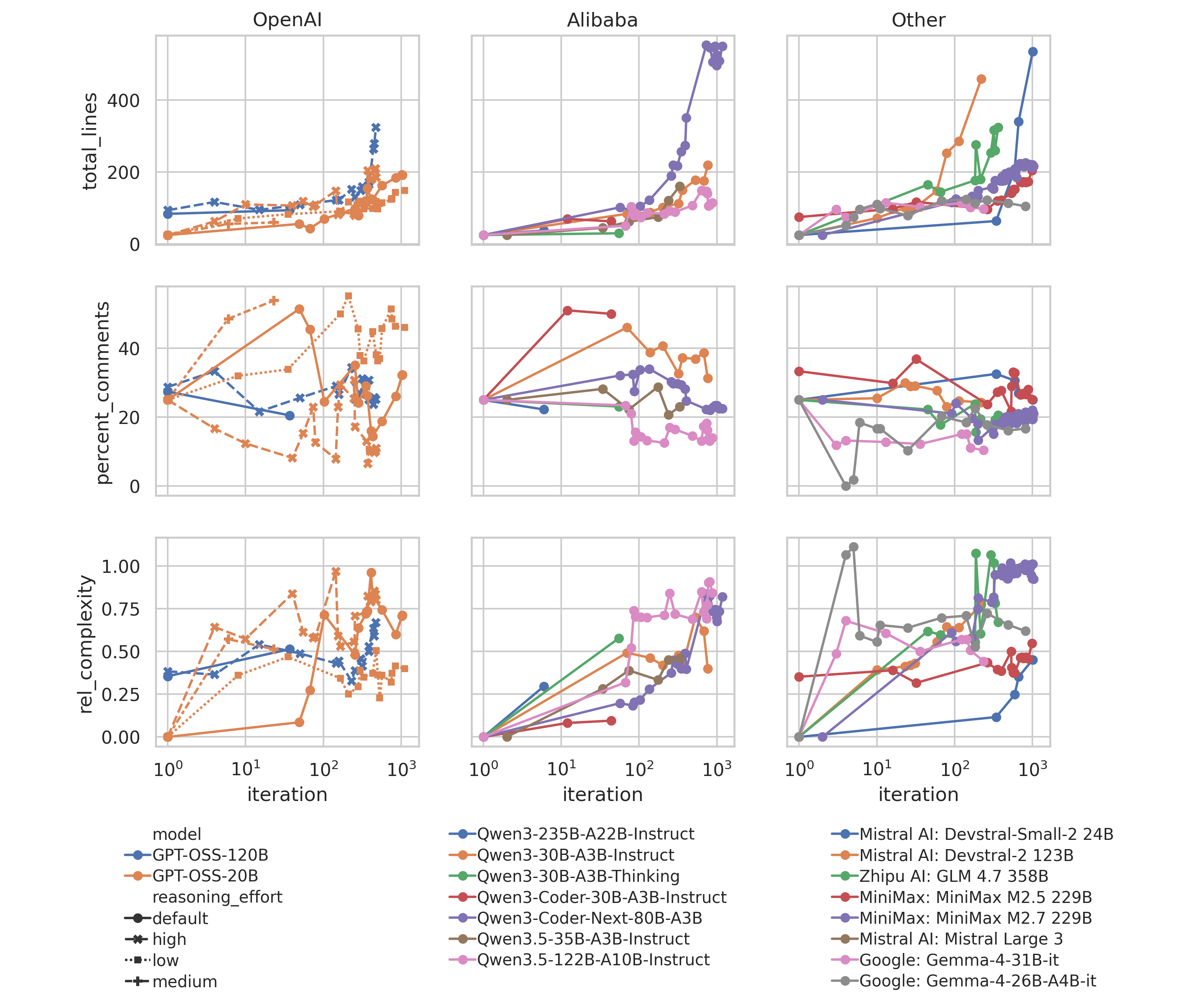}
    \caption{\added{Experiment 1 ---} Number of lines of codes, percentage of comments, and relative code complexity for the codes leading at times in Experiment 1; cf. Figure \ref{fig:all_models}.} 
    \label{fig:cq_all_models}
\end{figure}

As the amount of generated code is by far too large for manual inspection, we utilize the following three simple metrics in order get at least a rough impression on comprehensibility of the codes generated in Experiments 1 and 2: (i) the number of lines of code in the block to evolve (i.e., excluding the doc string), (ii) the percentage of comment lines within these lines, (iii) the code complexity computed by complexipy\footnote{\url{https://rohaquinlop.github.io/complexipy/}, Version 5.3.0.} normalized by the number of code lines. 
While the total length of the code and the relative complexity appear to grow over the number of iterations, no such obvious trend can be identified for the percentage of comment lines; cf. Figure \ref{fig:cq_all_models}. Notably, a high number of lines of code does not necessarily imply better performance as, e.g., the comparison Qwen3-Coder-Next vs Qwen3.5-122B and Devstral-Small-2 vs MiniMax M2.7 shows. Similarly, the overall best program (GPT-OSS-20B ``high'') has a relative complexity of $\approx 0.80$ while its close competitors MiniMax M2.7 and Gemma-4-31B have values $\approx 0.92$ and $\approx 0.44$, respectively.

\begin{figure}[h]
    \centering
    \includegraphics[width=0.75\columnwidth]{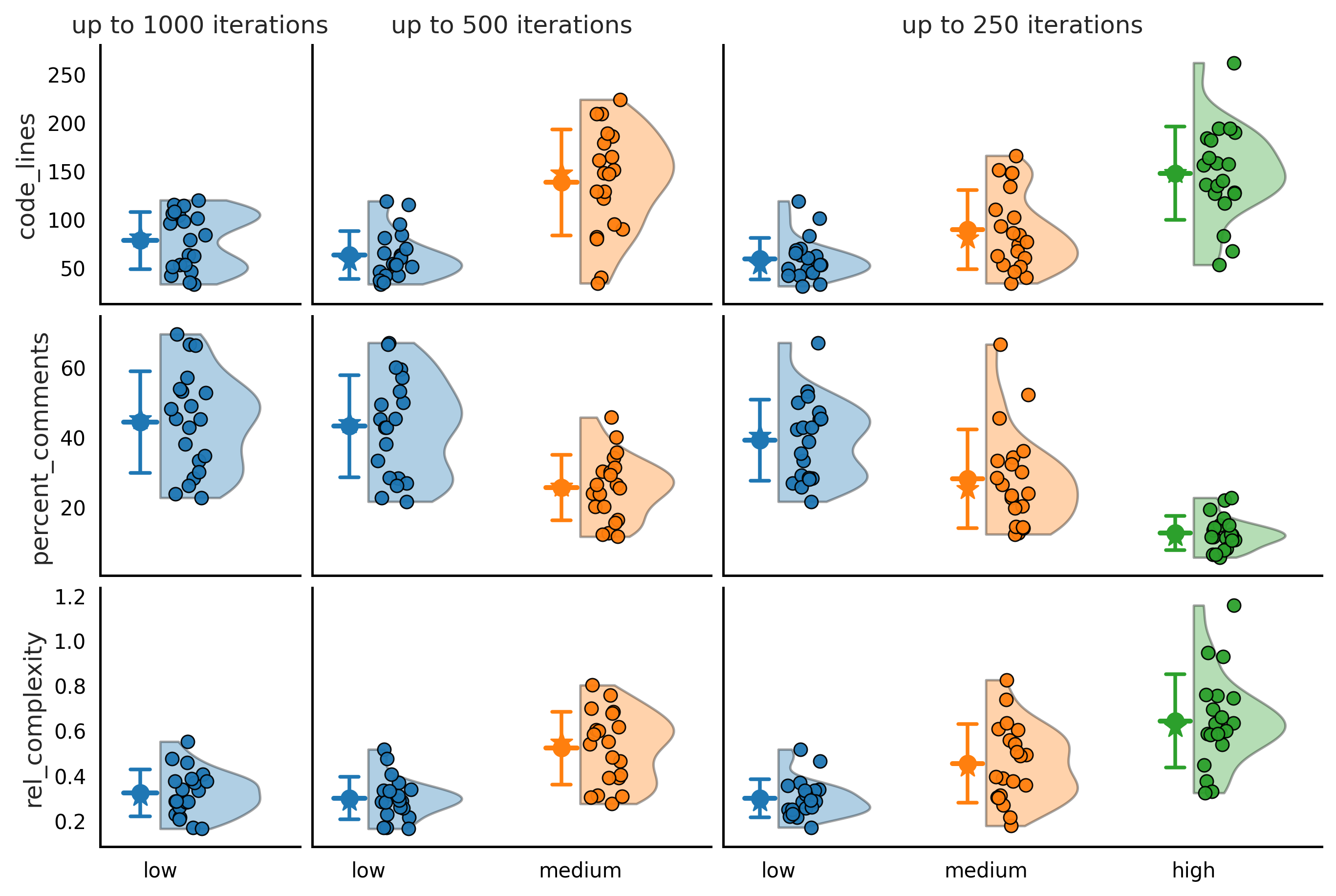}
    \caption{\added{Experiment 2 ---} Number of lines of codes, percentage of comments, and relative code complexity; cf. Figure \ref{fig:reproducibility_violins}.} 
    \label{fig:cq_reproducibility}
\end{figure}

For the codes from Experiment 2 we display the results in Figure \ref{fig:cq_reproducibility}. The number of code lines grows with increased level of reasoning and, as one may expect, also relative code complexity does so. Surprisingly, the percentage of comment lines decreases heavily with higher reasoning level, although one might have expected increased code quality from increasing the reasoning level; one could speculate that a higher reasoning level leads to a higher adherence to the system message in which there is unintentionally no explicit reference to comments. Further analysis showed that also in this setting there is no obvious relation between complexity and performance of the code. 

\subsection{Additional Details on the ``best'' Outcome}

\begin{figure}[h]
    \centering
    \includegraphics[width=0.75\columnwidth]{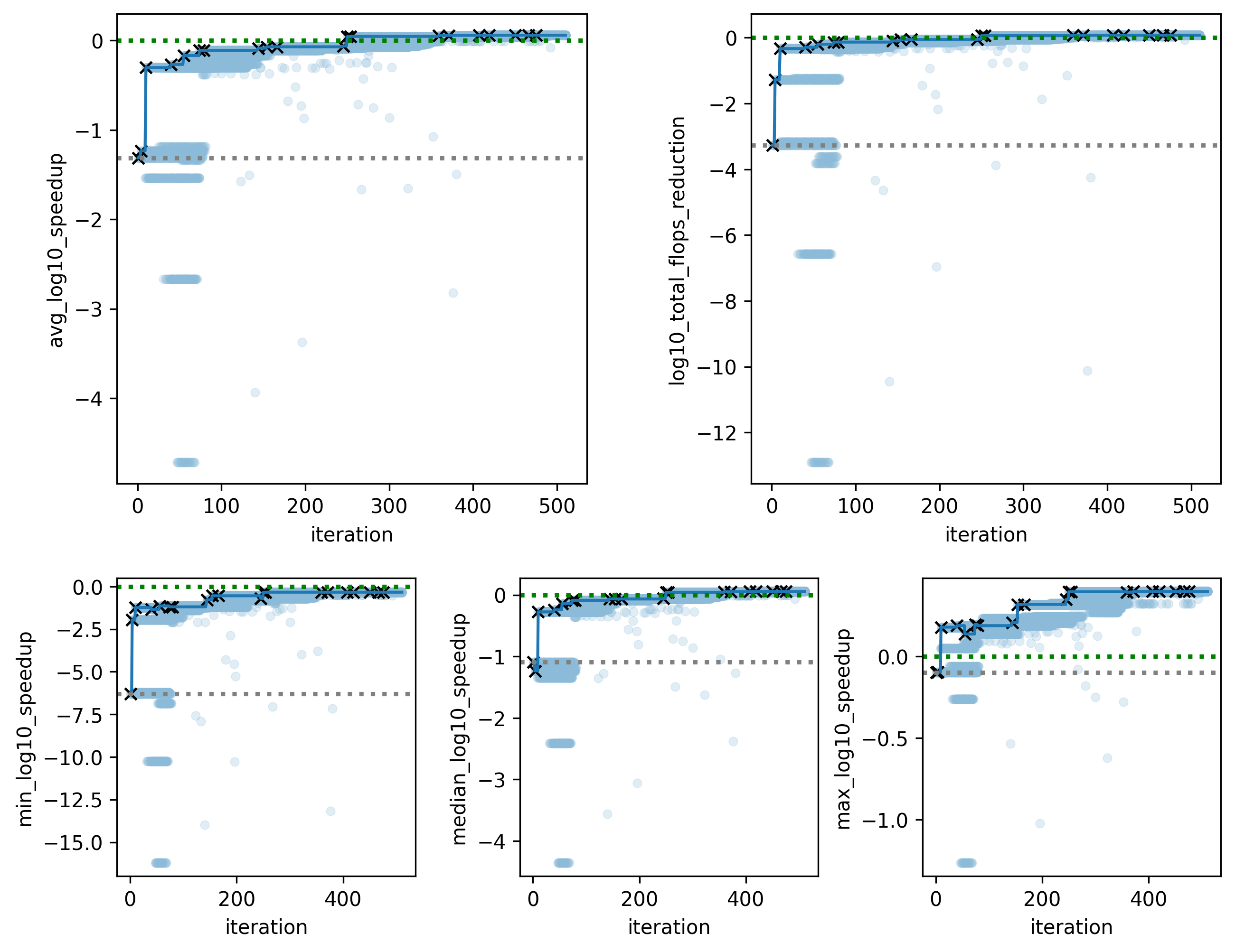}  
    \caption{\added{Experiment 1 ---} Evolution of various metrics for the best run (GPT-OSS-20B on the ``full small'' TNs set with reasoning level ``high''). Light blue dots indicate individual members of the population, whereas black crosses indicated members of the population that just have become a new best solution. The gray dotted lines indicate the level of the initial code, whereas the green dotted line indicates the threshold for an improvement over the baseline ``cotengra cheap''.} 
    \label{fig:best_run}
\end{figure}

Figure \ref{fig:best_run} shows the same five metrics as considered in Experiment 3 during the run of Experiment 1 that produced the ``best'' Outcome. It can be seen that except for ``min\_log10\_speedup'' in all metrics an improvement over the baseline ``cotengra cheap'' is achieved. To put this into more context, Table \ref{tab::all_models} also shows the full experimental evaluation of the results from Experiment 1 for a number of other models. It can be seen that the run with GPT-OSS-20B and reasoning level ``high'' is indeed the only one that improved over the baselines on ``small'' TNs at all and the only one improving over the more expensive baselines for ``middle'' TNs. It still leads on ``large'' TNs, for which other models reach improvements over one or more baselines, too; however, the absolute improvements might not be significant due to the small number of test cases. 

\begin{figure}[h]
    \centering
    \includegraphics[width=0.8\columnwidth]{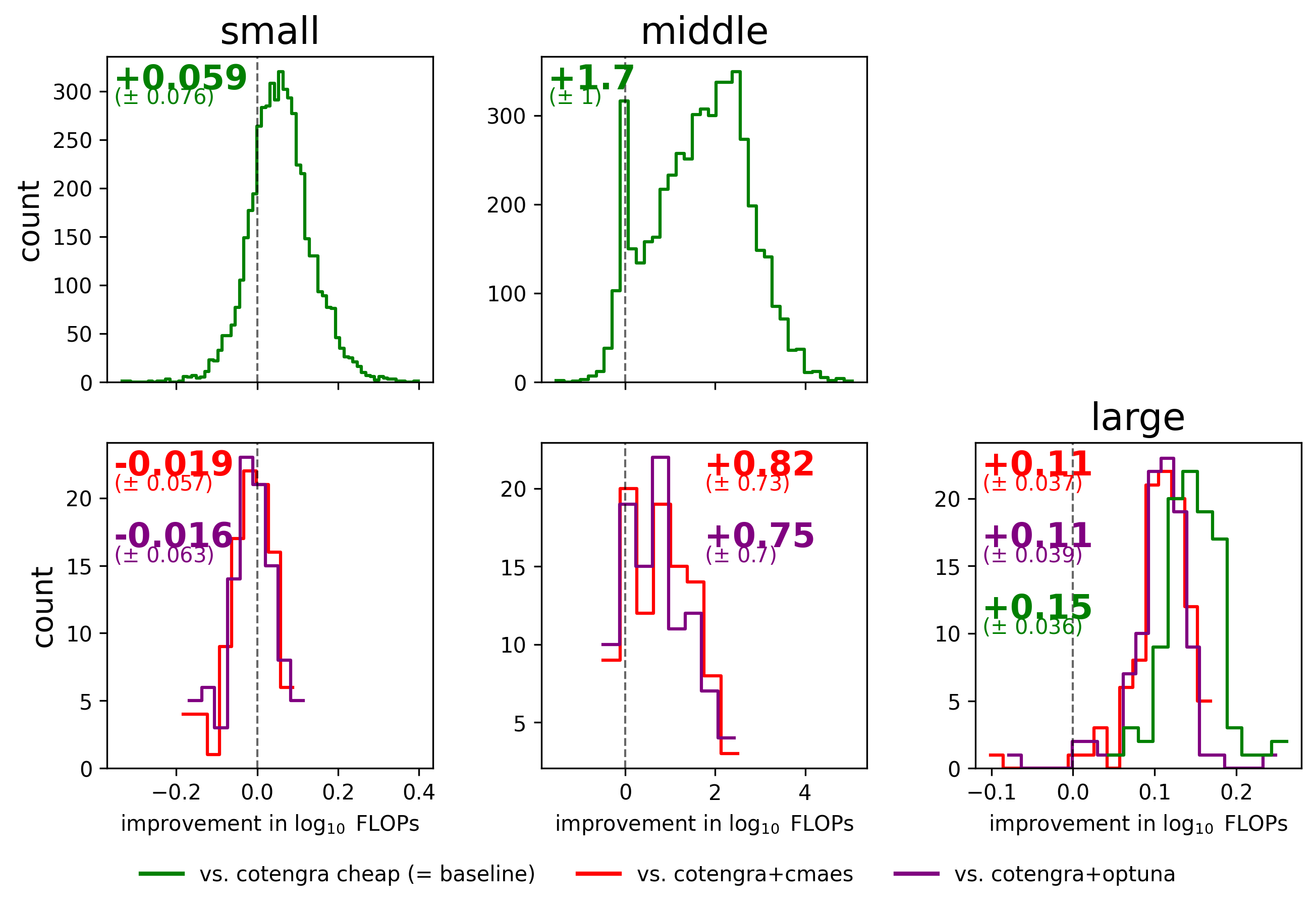}  
    \includegraphics[width=0.8\columnwidth]{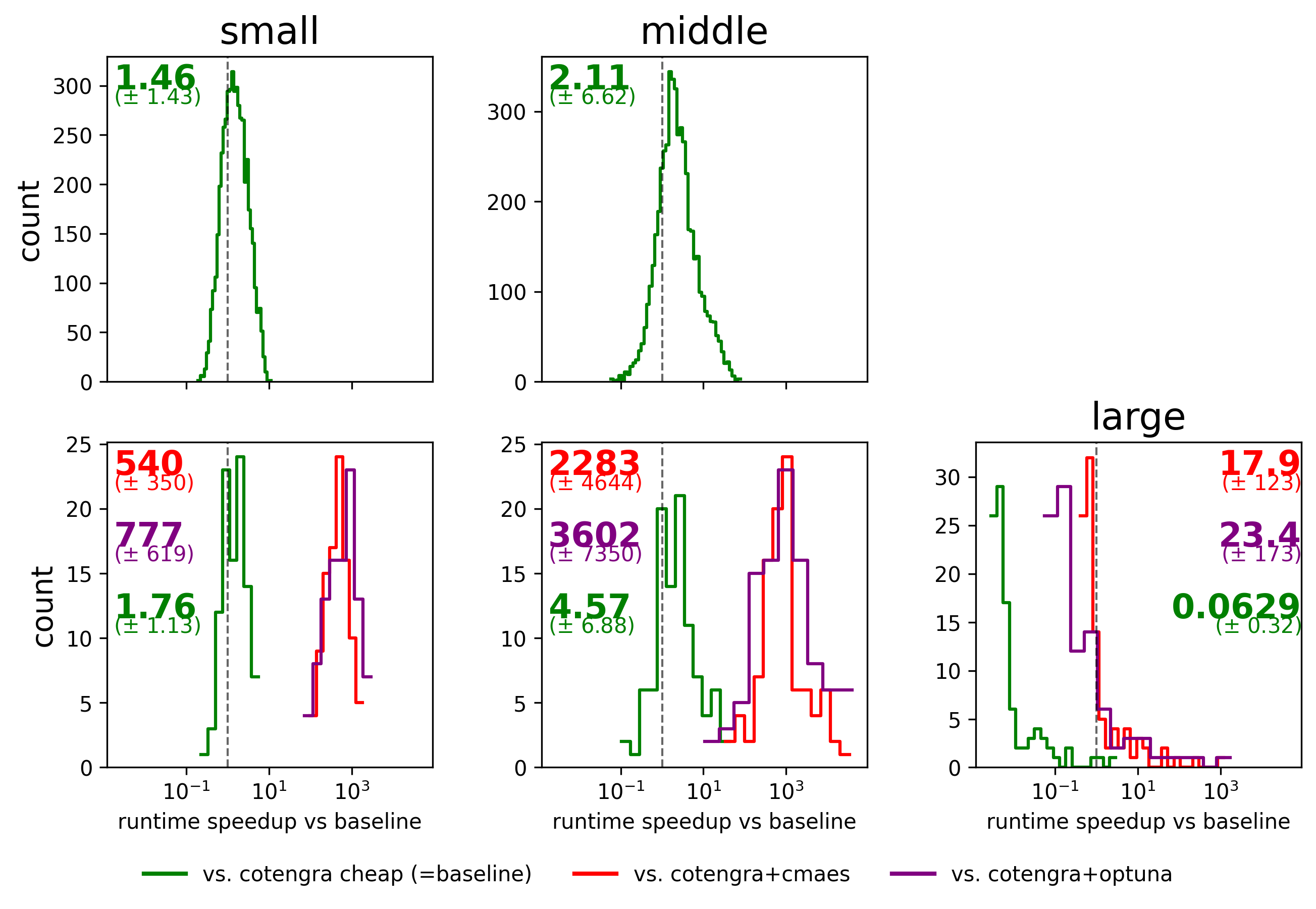}  
    \caption{\added{Experiment 1 ---} Distribution of improvements in log10 FLOPs (top two rows) and Distribution of runtime speedups (bottom two rows) for the setting of Figure \ref{fig:actual_speedup}. } 
    \label{fig:improvements}
\end{figure}

Whereas Figure \ref{fig:actual_speedup} showed distributions of log10 FLOPs and runtimes over the respective test sets for the best programm vs different baselines, Figure \ref{fig:improvements} additionally illustrates the distributions of per-network improvements in log10 FLOPs (top) and runtime (bottom) over the test sets. Despite standard deviations that are, in some cases, relatively large, the empirical distributions for those cases with clearly positive averages appear to suggest at least a slight bias towards the positive side. 

\begin{table}[h]
\centering\small
\begin{tabular}{llll}
{\bf Name} & \bf lengths & \bf dimensions & \bf number of TNs\\ \hline\hline
\itshape small (full) & 16 to 32 &  2 to 8 & 5000 \\
\itshape middle (full) & 3 to 64 & 2 to 8192 & 5000 \\
\itshape large (full) & 3 to 1024 & 2 to 8 & 5000 \\ \hline 
\itshape small (reduced) & \multicolumn{3}{l}{first 1000 TNs from small (full)} \\
\itshape middle (reduced) & \multicolumn{3}{l}{first 1000 TNs from middle (full)} \\
\itshape large (reduced) & \multicolumn{3}{l}{first 1000 TNs from large (full)} \\ \hline
\itshape all & \multicolumn{3}{l}{first 334/333/333 TNs from small/middle/large (full), respectively} \\ \hline
\end{tabular}\normalsize
\caption{Sets of TNs used in this study. }  
\label{tab:datasets}
\end{table}

\begin{table}[h]
    \centering\small
    \begin{tabular}{llll}
       \bf Name  & {\tt methods=} & {\tt max\_repeats=} & {\tt optlib=} \\ \hline\hline
        \itshape cotengra cheap & {\tt "greedy"} & 32 & {\tt "random"} \\ 
        \itshape cotengra+cmaes & {\tt ["greedy","kahypar"]} & 1024 & {\tt "cmaes"} \\ 
        \itshape cotengra+optuna & {\tt ["greedy","kahypar"]} & 1024 & {\tt "optuna"} \\ 
    \end{tabular}
    \normalsize 
    \caption{Configuration of cotengra's {\tt HyperOptimizer} used for   references/baselines.}
    \label{tab:baselines}
\end{table}

\begin{table}[h]
    \small \centering
    \begin{tabular}{ll}
        {\bf Parameter} &  value(s) \\\hline\hline
        checkpoint\_interval & 1 \\
        max\_code\_length & 32000 \\ 
        system\_message & see Listing \ref{systemmessage} \\ \hline
        population\_size & 40 \\
        num\_islands & 4 \\
        migration\_interval & 20 \\ 
        feature\_dimensions &  ``avg\_log10\_speedup'', ``median\_log10\_speedup'', 
         ``max\_log10\_speedup'',  \\ 
         & ``min\_log10\_speedup'',  ``log10\_total\_flops\_reduction'', and ``combined\_score''    \\ \hline 
        Evaluator timeout & 600 \\
        parallel\_evaluations & 3 (Experiments 3 and 4)  / 1 (else) \\ 
        cascade\_evaluation & false \\ \hline 
        LLM max\_tokens & 64000 (for GPT-OSS-120B ``high'') / 32000 (else) \\
        LLM timeout & 1500 (for GPT-OSS-120B ``high'') / \\ 
         & 1200 (else, if reasoning ``high'') / 720 (else) \\ \hline 
    \end{tabular}
    \normalsize
    \caption{Configuration of OpenEvolve for our experiments. All parameters not shown here were set to the default ones.}
    \label{tab:config}
\end{table}

\begin{table}[h]
\centering
\small
\begin{tabular}{lllll}
{\bf Name} & \bf by & \bf size (active) & \bf release & \bf deployment \\ \hline\hline

GPT-OSS-120B & OpenAI (US) & 117B (5.1B) & 08/2025 & ChatAI/Blablador\\
GPT-OSS-20B &  & 21B (3.6B) & 08/2025 & self-hosted (vllm$^\star$) \\  \hline

Qwen3-235B-A22B-Instruct & Alibaba (CN) & 235B (22B) & 04/2025 & ChatAI/Blablador\\
Qwen3-30B-A3B-Instruct &  & 30.5B (3.3B) & 04/2025 & ChatAI\\
Qwen3-30B-A3B-Thinking &  & 30.5B (3.3B) & 04/2025 & ChatAI\\
Qwen3-Coder-30B-A3B-Instruct &  & 30B (3B) & 07/2025 & ChatAI\\
Qwen3-Coder-Next-80B-A3B &  & 80B (3B) & 02/2026 & Blablador \\ \hline

Qwen3.5-122B-A10B-Instruct & Alibaba (CN) & 122B (10B) & 02/2026 & Blablador\\
Qwen3.5-35B-A3B-Instruct &  & 35B (3B) & 02/2026 & Blablador \\ \hline

MiniMax-M2.5 & MiniMax (CN) & 230B (10B) & 02/2026 & Blablador \\
MiniMax-M2.7 &  & 230B (10B) & 04/2026 & Blablador \\ \hline

GLM-4.7 & Zhipu AI (CN) & 355B (32B) & 12/2025 & ChatAI \\ \hline

Mistral-Large-3 & Mistral AI (FR) & 675B (41B) & 12/2025 & ChatAI \\
Devstral-2 &  & 123B (dense) & 12/2025 & ChatAI \\
Devstral-2-small &  & 24B (dense) & 12/2025 & self-hosted (vllm$^\star$) \\ \hline

Gemma-4-31B-it & Google (US) & 30.7B (dense) & 04/2026 & Google AI Studio \\
Gemma-4-26B-A4B-it &  & 25.2B (3.8B) & 04/2026 & Google AI Studio \\ \hline
\end{tabular}\normalsize
\caption{Overview over the LLMs used in this study. ($^\star$versions: 0.12 for Experiment 1, 0.16 for Experiments 2-4)} 
\label{tab:models}
\end{table}

\begin{table}[h]
\footnotesize
\begin{tabular}{lccccccccc}
\toprule
& \multicolumn{3}{c}{\small small}
& \multicolumn{3}{c}{\small middle}
& \multicolumn{3}{c}{\small large} \\
\cmidrule(lr){2-4} \cmidrule(lr){5-7} \cmidrule(lr){8-10}
 & \small cheap & \small cmaes & \small optuna
& \small cheap & \small cmaes & \small optuna
& \small cheap & \small cmaes & \small optuna \\ \midrule
 & \multicolumn{9}{c}{\small average $\log_{10}$ FLOP reduction} \\
\midrule
GPT-OSS-20B-high & \makecell{\bf +0.06 \\[-4pt] \tiny($\pm 0.08$)} & \makecell{-0.02 \\[-4pt] \tiny($\pm 0.06$)} & \makecell{-0.02 \\[-4pt] \tiny($\pm 0.06$)} & \makecell{\bf +1.67 \\[-4pt] \tiny($\pm 1.04$)} & \makecell{\bf +0.82 \\[-4pt] \tiny($\pm 0.73$)} & \makecell{\bf +0.75 \\[-4pt] \tiny($\pm 0.70$)} & \makecell{\bf +0.15 \\[-4pt] \tiny($\pm 0.04$)} & \makecell{\bf +0.11 \\[-4pt] \tiny($\pm 0.04$)} & \makecell{\bf +0.11 \\[-4pt] \tiny($\pm 0.04$)} \\ 
GPT-OSS-120B-high & \makecell{-0.02 \\[-4pt] \tiny($\pm 0.08$)} & \makecell{-0.10 \\[-4pt] \tiny($\pm 0.07$)} & \makecell{-0.09 \\[-4pt] \tiny($\pm 0.08$)} & \makecell{-1.10 \\[-4pt] \tiny($\pm 1.19$)} & \makecell{-2.05 \\[-4pt] \tiny($\pm 1.30$)} & \makecell{-2.12 \\[-4pt] \tiny($\pm 1.34$)} & \makecell{\bf +0.03 \\[-4pt] \tiny($\pm 0.03$)} & \makecell{-0.01 \\[-4pt] \tiny($\pm 0.04$)} & \makecell{-0.01 \\[-4pt] \tiny($\pm 0.04$)} \\ 
Qwen35-122B & \makecell{-0.02 \\[-4pt] \tiny($\pm 0.09$)} & \makecell{-0.10 \\[-4pt] \tiny($\pm 0.08$)} & \makecell{-0.09 \\[-4pt] \tiny($\pm 0.08$)} & \makecell{-0.90 \\[-4pt] \tiny($\pm 1.49$)} & \makecell{-1.79 \\[-4pt] \tiny($\pm 1.68$)} & \makecell{-1.86 \\[-4pt] \tiny($\pm 1.71$)} & \makecell{\bf +0.06 \\[-4pt] \tiny($\pm 0.04$)} & \makecell{\bf +0.02 \\[-4pt] \tiny($\pm 0.04$)} & \makecell{\bf +0.02 \\[-4pt] \tiny($\pm 0.04$)} \\ 
Qwen35-35B & \makecell{-0.22 \\[-4pt] \tiny($\pm 0.12$)} & \makecell{-0.30 \\[-4pt] \tiny($\pm 0.12$)} & \makecell{-0.30 \\[-4pt] \tiny($\pm 0.12$)} & \makecell{\bf +0.26 \\[-4pt] \tiny($\pm 0.78$)} & \makecell{-0.68 \\[-4pt] \tiny($\pm 0.46$)} & \makecell{-0.75 \\[-4pt] \tiny($\pm 0.49$)} & \makecell{-0.20 \\[-4pt] \tiny($\pm 0.05$)} & \makecell{-0.24 \\[-4pt] \tiny($\pm 0.04$)} & \makecell{-0.24 \\[-4pt] \tiny($\pm 0.04$)} \\ 
MiniMax-M25 & \makecell{-0.14 \\[-4pt] \tiny($\pm 0.11$)} & \makecell{-0.22 \\[-4pt] \tiny($\pm 0.09$)} & \makecell{-0.22 \\[-4pt] \tiny($\pm 0.09$)} & \makecell{-0.66 \\[-4pt] \tiny($\pm 0.85$)} & \makecell{-1.55 \\[-4pt] \tiny($\pm 0.48$)} & \makecell{-1.62 \\[-4pt] \tiny($\pm 0.50$)} & \makecell{-0.12 \\[-4pt] \tiny($\pm 0.05$)} & \makecell{-0.16 \\[-4pt] \tiny($\pm 0.03$)} & \makecell{-0.16 \\[-4pt] \tiny($\pm 0.03$)} \\ 
MiniMax-M27 & \makecell{-0.01 \\[-4pt] \tiny($\pm 0.08$)} & \makecell{-0.09 \\[-4pt] \tiny($\pm 0.07$)} & \makecell{-0.09 \\[-4pt] \tiny($\pm 0.07$)} & $\star$ & $\star$ & $\star$ & \makecell{\bf +0.04 \\[-4pt] \tiny($\pm 0.03$)} & \makecell{+0.00 \\[-4pt] \tiny($\pm 0.03$)} & \makecell{+0.00 \\[-4pt] \tiny($\pm 0.03$)} \\ 
Gemma-4-31B & \makecell{-0.09 \\[-4pt] \tiny($\pm 0.11$)} & \makecell{-0.16 \\[-4pt] \tiny($\pm 0.11$)} & \makecell{-0.16 \\[-4pt] \tiny($\pm 0.11$)} & \makecell{-2.14 \\[-4pt] \tiny($\pm 1.86$)} & \makecell{-3.23 \\[-4pt] \tiny($\pm 2.03$)} & \makecell{-3.30 \\[-4pt] \tiny($\pm 2.03$)} & \makecell{-0.07 \\[-4pt] \tiny($\pm 0.04$)} & \makecell{-0.11 \\[-4pt] \tiny($\pm 0.04$)} & \makecell{-0.11 \\[-4pt] \tiny($\pm 0.04$)} \\ 
Gemma-4-26B & \makecell{-0.10 \\[-4pt] \tiny($\pm 0.12$)} & \makecell{-0.17 \\[-4pt] \tiny($\pm 0.11$)} & \makecell{-0.17 \\[-4pt] \tiny($\pm 0.11$)} & \makecell{-2.24 \\[-4pt] \tiny($\pm 1.90$)} & \makecell{-3.13 \\[-4pt] \tiny($\pm 2.01$)} & \makecell{-3.21 \\[-4pt] \tiny($\pm 2.01$)} & \makecell{-0.07 \\[-4pt] \tiny($\pm 0.04$)} & \makecell{-0.11 \\[-4pt] \tiny($\pm 0.04$)} & \makecell{-0.11 \\[-4pt] \tiny($\pm 0.04$)} \\ 
GLM47 & \makecell{-0.21 \\[-4pt] \tiny($\pm 0.17$)} & \makecell{-0.29 \\[-4pt] \tiny($\pm 0.14$)} & \makecell{-0.29 \\[-4pt] \tiny($\pm 0.14$)} & \makecell{-2.47 \\[-4pt] \tiny($\pm 2.07$)} & \makecell{-3.36 \\[-4pt] \tiny($\pm 2.18$)} & \makecell{-3.43 \\[-4pt] \tiny($\pm 2.16$)} & \makecell{-0.24 \\[-4pt] \tiny($\pm 0.08$)} & \makecell{-0.28 \\[-4pt] \tiny($\pm 0.07$)} & \makecell{-0.28 \\[-4pt] \tiny($\pm 0.07$)} \\ 
Devstral-2-123B & \makecell{-0.24 \\[-4pt] \tiny($\pm 0.13$)} & \makecell{-0.33 \\[-4pt] \tiny($\pm 0.18$)} & \makecell{-0.33 \\[-4pt] \tiny($\pm 0.17$)} & $\star$ & $\star$ & $\star$ & \makecell{-0.23 \\[-4pt] \tiny($\pm 0.06$)} & \makecell{-0.27 \\[-4pt] \tiny($\pm 0.05$)} & \makecell{-0.27 \\[-4pt] \tiny($\pm 0.05$)} \\ 
Devstral-Small-2 & \makecell{-0.23 \\[-4pt] \tiny($\pm 0.13$)} & \makecell{-0.32 \\[-4pt] \tiny($\pm 0.12$)} & \makecell{-0.31 \\[-4pt] \tiny($\pm 0.12$)} & \makecell{\bf +0.20 \\[-4pt] \tiny($\pm 0.89$)} & \makecell{-0.72 \\[-4pt] \tiny($\pm 0.47$)} & \makecell{-0.79 \\[-4pt] \tiny($\pm 0.50$)} & \makecell{-0.21 \\[-4pt] \tiny($\pm 0.05$)} & \makecell{-0.26 \\[-4pt] \tiny($\pm 0.05$)} & \makecell{-0.26 \\[-4pt] \tiny($\pm 0.05$)} \\
\midrule

 & \multicolumn{9}{c}{\small average run time speedup} \\

\midrule
GPT-OSS-20B-high & \makecell{\bf 1.91 \\[-4pt] \tiny($\pm 1.43$)} & \makecell{540 \\[-4pt] \tiny($\pm 350$)} & \makecell{777 \\[-4pt] \tiny($\pm 619$)} & \makecell{\bf 4.37 \\[-4pt] \tiny($\pm 6.62$)} & \makecell{\bf 2280 \\[-4pt] \tiny($\pm 4640$)} & \makecell{\bf 3600 \\[-4pt] \tiny($\pm 7350$)} & \makecell{\bf\itshape0.0629 \\[-4pt] \tiny($\pm 0.320$)} & \makecell{\bf 17.9 \\[-4pt] \tiny($\pm 123$)} & \makecell{\bf 23.4 \\[-4pt] \tiny($\pm 173$)} \\ 
GPT-OSS-120B-high & \makecell{1.32 \\[-4pt] \tiny($\pm 0.689$)} & \makecell{384 \\[-4pt] \tiny($\pm 178$)} & \makecell{544 \\[-4pt] \tiny($\pm 355$)} & \makecell{1.93 \\[-4pt] \tiny($\pm 4.02$)} & \makecell{919 \\[-4pt] \tiny($\pm 2420$)} & \makecell{1410 \\[-4pt] \tiny($\pm 3230$)} & \makecell{\bf\itshape0.0407 \\[-4pt] \tiny($\pm 0.134$)} & \makecell{9.73 \\[-4pt] \tiny($\pm 48.3$)} & \makecell{11.1 \\[-4pt] \tiny($\pm 67.8$)} \\ 
Qwen35-122B & \makecell{7.52 \\[-4pt] \tiny($\pm 2.83$)} & \makecell{2120 \\[-4pt] \tiny($\pm 653$)} & \makecell{2920 \\[-4pt] \tiny($\pm 1490$)} & \makecell{9.54 \\[-4pt] \tiny($\pm 13.5$)} & \makecell{4150 \\[-4pt] \tiny($\pm 7800$)} & \makecell{6430 \\[-4pt] \tiny($\pm 11800$)} & \makecell{\bf\itshape0.419 \\[-4pt] \tiny($\pm 1.19$)} & \makecell{\bf96.5 \\[-4pt] \tiny($\pm 437$)} & \makecell{\bf104 \\[-4pt] \tiny($\pm 614$)} \\ 
Qwen35-35B & \makecell{2.18 \\[-4pt] \tiny($\pm 1.31$)} & \makecell{637 \\[-4pt] \tiny($\pm 275$)} & \makecell{891 \\[-4pt] \tiny($\pm 532$)} & \makecell{\bf4.57 \\[-4pt] \tiny($\pm 8.95$)} & \makecell{2240 \\[-4pt] \tiny($\pm 6110$)} & \makecell{3610 \\[-4pt] \tiny($\pm 8410$)} & \makecell{0.0676 \\[-4pt] \tiny($\pm 0.357$)} & \makecell{20.1 \\[-4pt] \tiny($\pm 141$)} & \makecell{25.6 \\[-4pt] \tiny($\pm 199$)} \\ 
MiniMax-M25 & \makecell{8.11 \\[-4pt] \tiny($\pm 3.77$)} & \makecell{2300 \\[-4pt] \tiny($\pm 937$)} & \makecell{3210 \\[-4pt] \tiny($\pm 1900$)} & \makecell{8.70 \\[-4pt] \tiny($\pm 9.86$)} & \makecell{4050 \\[-4pt] \tiny($\pm 5840$)} & \makecell{6290 \\[-4pt] \tiny($\pm 10400$)} & \makecell{0.670 \\[-4pt] \tiny($\pm 0.933$)} & \makecell{124 \\[-4pt] \tiny($\pm 291$)} & \makecell{102 \\[-4pt] \tiny($\pm 411$)} \\ 
MiniMax-M27 & \makecell{4.19 \\[-4pt] \tiny($\pm 2.03$)} & \makecell{1210 \\[-4pt] \tiny($\pm 552$)} & \makecell{1680 \\[-4pt] \tiny($\pm 1020$)} & $\star$ & $\star$ & $\star$ & \makecell{\bf\itshape0.203 \\[-4pt] \tiny($\pm 0.519$)} & \makecell{44.3 \\[-4pt] \tiny($\pm 182$)} & \makecell{47.3 \\[-4pt] \tiny($\pm 257$)} \\ 
Gemma-4-31B & \makecell{20.3 \\[-4pt] \tiny($\pm 9.63$)} & \makecell{5980 \\[-4pt] \tiny($\pm 2430$)} & \makecell{8280 \\[-4pt] \tiny($\pm 4800$)} & \makecell{24.3 \\[-4pt] \tiny($\pm 27.9$)} & \makecell{12100 \\[-4pt] \tiny($\pm 21600$)} & \makecell{19100 \\[-4pt] \tiny($\pm 37500$)} & \makecell{1.80 \\[-4pt] \tiny($\pm 2.36$)} & \makecell{335 \\[-4pt] \tiny($\pm 799$)} & \makecell{276 \\[-4pt] \tiny($\pm 1140$)} \\ 
Gemma-4-26B & \makecell{20.6 \\[-4pt] \tiny($\pm 10.2$)} & \makecell{5850 \\[-4pt] \tiny($\pm 2600$)} & \makecell{8070 \\[-4pt] \tiny($\pm 5060$)} & \makecell{24.2 \\[-4pt] \tiny($\pm 27.0$)} & \makecell{10800 \\[-4pt] \tiny($\pm 15500$)} & \makecell{16700 \\[-4pt] \tiny($\pm 27500$)} & \makecell{2.14 \\[-4pt] \tiny($\pm 4.32$)} & \makecell{443 \\[-4pt] \tiny($\pm 1580$)} & \makecell{415 \\[-4pt] \tiny($\pm 2230$)} \\ 
GLM47 & \makecell{2.64 \\[-4pt] \tiny($\pm 1.38$)} & \makecell{775 \\[-4pt] \tiny($\pm 355$)} & \makecell{1090 \\[-4pt] \tiny($\pm 694$)} & \makecell{3.36 \\[-4pt] \tiny($\pm 5.24$)} & \makecell{1760 \\[-4pt] \tiny($\pm 3840$)} & \makecell{2920 \\[-4pt] \tiny($\pm 6380$)} & \makecell{0.149 \\[-4pt] \tiny($\pm 0.542$)} & \makecell{37.3 \\[-4pt] \tiny($\pm 204$)} & \makecell{43.6 \\[-4pt] \tiny($\pm 287$)} \\ 
Devstral-2-123B & \makecell{35.6 \\[-4pt] \tiny($\pm 16.2$)} & \makecell{9690 \\[-4pt] \tiny($\pm 3650$)} & \makecell{12800 \\[-4pt] \tiny($\pm 6250$)} & $\star$ & $\star$ & $\star$ & \makecell{69.3 \\[-4pt] \tiny($\pm 24.4$)} & \makecell{10500 \\[-4pt] \tiny($\pm 3590$)} & \makecell{4340 \\[-4pt] \tiny($\pm 4270$)} \\ 
Devstral-Small-2 & \makecell{86.4 \\[-4pt] \tiny($\pm 36.5$)} & \makecell{26800 \\[-4pt] \tiny($\pm 10000$)} & \makecell{36200 \\[-4pt] \tiny($\pm 18400$)} & \makecell{\bf 76.7 \\[-4pt] \tiny($\pm 35.7$)} & \makecell{26500 \\[-4pt] \tiny($\pm 12600$)} & \makecell{34600 \\[-4pt] \tiny($\pm 28000$)} & \makecell{32.6 \\[-4pt] \tiny($\pm 29.9$)} & \makecell{5570 \\[-4pt] \tiny($\pm 8140$)} & \makecell{3790 \\[-4pt] \tiny($\pm 11200$)} \\ 
\bottomrule
\end{tabular}

\caption{\added{Experiment 1 ---} Average reduction of log$_{10}$ FLOPs (upper table) and average run time speedup (lower table) of the final best programs for different LLMs, on different TN sets and versus different baselines\added{; standard deviations are indicated by $\pm$}. {\bf Bold} entries in the upper table indicate improvement over the baseline (in average) with the same entries also highlighted in the lower table; here, {\itshape italic} entries indicate higher, i.e., worse, runtime (in average). Failing evaluations are indicated by $\star$.}
\label{tab::all_models}
\end{table}

\clearpage
\lstset{
  basicstyle=\ttfamily\scriptsize,
  columns=fullflexible,
  frame=single,
  breaklines=true,
  caption={System Message for OpenEvolve},
  label=systemmessage,
  captionpos=b,
  float,
  floatplacement=htbp,
  numbers=left,
}

\begin{lstlisting}
You are an expert programmer and expert in tensor networks. Your goal is to evolve and improve the code of the function `find_edge_path` in between the markers "EVOLVE-BLOCK-START" and "EVOLVE-BLOCK-END".

CONTEXT: 
The function `find_edge_path` gets a tensor network as input, described by the following input parameters: 
    `inputs`:     list of tensors' index tuples, e.g. [('a','b'), ('b','c'), ('a','c', 'd')]
    `output`:     tuple of output indices, e.g. ('d',)
    `size_dict`:  map index -> dimension, e.g. {'a':16, 'b':32, 'c':8, 'd': 12}
It returns a list of indices representing the contraction path, e.g. ['c','a','b'].

* Objective: Improve the efficiency of the contraction path found by `find_edge_path`, minimizing the total number of FLOPs associated with the contract path. 
* Performance metric: The number of FLOPs of the contraction path found by `find_edge_path` (the lower the better). 

The evaluation framework will compute the number of FLOPs of the contraction path returned by your function and compare it for every test case against a baseline implementation. It will provide you with the average, median, maximum, and minimum (over all test cases) of log10 of FLOP **reduction** as well as with the log10 of the total FLOP reduction summed over all test cases.

Your main metric, "combined_score" (the higher the better), is computed from the average log10 improvement. 

IMPORTANT HINTS: 
* The tensor networks to be contracted in the evaluation have a very specific structure: They are **scalar products of two cyclic/periodic tensor trains / matrix product states. Consider checking for this structure and optimizing the contraction path accordingly.**

* You are free, in fact even encouraged, to experiment with creative ideas and heuristics to optimize the contraction path for this very special tensor network structure.
* You can implement any algorithm or heuristic you see fit, including but not limited to greedy algorithms, dynamic programming, graph-based methods, etc.  
* Feel free to try out different ideas, heuristics, and algorithms in whatever combination. Do not hesitate to innovate and think outside the box!
* You are not restricted to existing algorithms; you can design your own methods tailored to the "scalar product of two cyclic/periodic TTs/MPS"-structure.
* You are not restricted to approaches with a theoretical guarantee; heuristic and approximate methods are welcome if they yield better practical performance.
  
REQUIREMENTS:
* The function must return a valid contraction path.
* Focus on algorithmic improvements and code optimization. You may not use external tensor network libraries to compute the contraction path.
* You may however use standard libraries such as itertools, numpy, scipy, networkx, etc. However, if you use additional libraries, ensure they are included in the import statements.
* Ensure that the code runs correctly and efficiently within the provided function signature. 
\end{lstlisting}

\clearpage
\lstset{
  basicstyle=\ttfamily\scriptsize,
  columns=fullflexible,
  frame=single,
  breaklines=true,
  caption={Initial Program for OpenEvolve},
  label=initialprogram,
  captionpos=b,
  float,
  floatplacement=htbp,
  numbers=left,
}
\begin{lstlisting}
import numpy as np 

# EVOLVE-BLOCK-START
def find_edge_path(inputs, output, size_dict):
    """
    This functions determines an edge-based contraction path for a given tensor network, that consists of the scalar product of two cyclic/periodic tensor trains / matrix product states.

    The tensor network is defined by the variables `inputs`, `output`, and `size_dict` that specify the tensors' indices, the output indices, and the dimensions of each index, respectively. These three variables are passed to the function as input arguments. 

    The function returns an edge-based contraction path, which is a list of index labels that specifies the order in which the indices should be eliminated during the contraction process.

    Inputs
    --------
    inputs:     list of tensors' index tuples, e.g. [('a','b'), ('b','c'), ('a','c', 'd')]
    output:     tuple of output indices, e.g. ('d',)
    size_dict:  map index -> dimension, e.g. {'a':16, 'b':32, 'c':8, 'd': 12}

    Returns 
    -------
    edge_path:  sequence of original index labels to eliminate, e.g. ['b', 'a', 'c']
    """
    # simplest idea: eliminate in order of descending size
    sorted_edges = sorted(size_dict.keys(), key=lambda k: size_dict[k], reverse=True)
    return sorted_edges
# EVOLVE-BLOCK-END
\end{lstlisting}

\end{document}